# Multilingual Text Style Transfer: Datasets & Models for Indian Languages

**Sourabrata Mukherjee**[1], **Atul Kr. Ojha**[2,3], **Akanksha Bansal**[3], **Deepak Alok**[3]
**John P. McCrae**[2], **Ondřej Dušek**[1]
[1]Charles University, Faculty of Mathematics and Physics, Prague, Czechia
[2]Insight SFI Research Centre for Data Analytics, DSI, University of Galway, Ireland
[3]Panlingua Language Processing LLP, India
{mukherjee,odusek}@ufal.mff.cuni.cz
{akanksha.bansal,deepak.alok}@panlingua.co.in
{atulkumar.ojha,john.mccrae}@insight-centre.org

## Abstract

Text style transfer (TST) involves altering the linguistic style of a text while preserving its style-independent content. This paper focuses on sentiment transfer, a popular TST subtask, across a spectrum of Indian languages: Hindi, Magahi, Malayalam, Marathi, Punjabi, Odia, Telugu, and Urdu, expanding upon previous work on English-Bangla sentiment transfer (Mukherjee et al., 2023a). We introduce dedicated datasets of 1,000 positive and 1,000 negative style-parallel sentences for each of these eight languages. We then evaluate the performance of various benchmark models categorized into parallel, non-parallel, cross-lingual, and shared learning approaches, including the Llama2 and GPT-3.5 large language models (LLMs). Our experiments highlight the significance of parallel data in TST and demonstrate the effectiveness of the Masked Style Filling (MSF) approach (Mukherjee et al., 2023a) in non-parallel techniques. Moreover, cross-lingual and joint multilingual learning methods show promise, offering insights into selecting optimal models tailored to the specific language and task requirements. To the best of our knowledge, this work represents the first comprehensive exploration of the TST task as sentiment transfer across a diverse set of languages.

## 1 Introduction

Text Style Transfer (TST) is an evolving field within natural language processing that has gained prominence for its capacity to modify the style of a given text while preserving its fundamental content (Mukherjee and Dušek, 2024; Mukherjee et al., 2024a). Notably, TST has primarily been explored in English, leaving a significant gap in linguistic diversity and a lack of comprehensive resources for effective multilingual style transfer. This research aims to bridge this gap by extending the boundaries of TST to include other diverse Indian languages: Hindi, Magahi, Malayalam, Marathi, Punjabi, Odia, Telugu, and Urdu.

We work with sentiment transfer and use the English dataset of Mukherjee et al. (2023a), who experimented with English and Bangla. We have extended the scope by adding eight new languages to the dataset. We manually translated the English dataset into other languages to maintain the style, content, and structural alignment, prioritizing naturalness in the target language (details in Section 3.2). We created new multilingual TST datasets using human annotators. They serve as counterparts to the refined English dataset (Mukherjee et al., 2023a) in a well-established linguistic context.

In addition, we tested several standard models (see Section 4) to validate and assess the quality and usefulness of the language-specific datasets.

Our contributions are summarized as follows:

(i) We introduce new multilingual datasets for sentiment transfer that align with the English counterpart, expanding the resources for TST tasks across multiple languages.

(ii) Using our datasets, we conducted experiments using multiple previously proposed models for TST as well as LLMs (Mukherjee et al., 2024b), including a scenario with no parallel data and the use of machine translation. We also include joint multilingual training, leveraging information exchange across languages for improved TST task performance.

(iii) We provide a detailed analysis of the results-facilitating a comprehensive understanding of the multi-lingual cross-linguistic effectiveness of our approaches.

(iv) Our data and experimental code are released on GitHub.[1]

[1]Code: https://github.com/souro/multilingual_tst, data: https://github.com/panlingua/multilingual-tst-datasets.

## 2 Related Work

TST typically involves training on pairs of texts that share content but differ in style. For example, Jhamtani et al. (2017) used a sequence-to-sequence model with a pointer network to transform modern English into Shakespearean English. Meanwhile, Mukherjee and Dusek (2023) employed minimal parallel data and integrated various low-resource methods for TST. However, this approach is particularly challenging due to the limited availability of parallel data (Hu et al., 2022; Mukherjee et al., 2023a).

To reduce the need for parallel data, two main strategies have been used: (i) Simple text replacement, where specific style-related phrases are explicitly identified and substituted (Li et al., 2018; Mukherjee et al., 2023a). (ii) Implicitly disentangling style from content through latent representations, using techniques like back-translation and autoencoding (Mukherjee et al., 2022; Zhao et al., 2018; Fu et al., 2018; Prabhumoye et al., 2018a; Hu et al., 2017). However, non-parallel approaches often produce mixed results and require significant amounts of stylized non-parallel data, which can be scarce for many styles (Mukherjee et al., 2022; Li et al., 2022).

In our experiments in Section 4, we evaluate both approaches using low-resource parallel data and non-parallel approaches.

**Multilingual style transfer** is a relatively unexplored area in prior research. Briakou et al. (2021) presented a multilingual formality style transfer benchmark, XFORMAL, including languages like Chinese, Russian, Latvian, Estonian, and French. Moreover, Krishna et al. (2022) focused on altering formality in various Indian languages. To the best of our knowledge, we are the first to explore text sentiment transfer within the domain of TST for the languages under consideration. We follow both above works by evaluating models on our benchmark in multilingual as well as crosslingual setups.

## 3 Dataset Preparation

We decided to base our effort on the Yelp dataset of Mukherjee et al. (2023a), as it offered a suitable size, parallel structure, and a relevant domain for our efforts. The dataset consists of 1,000 style-parallel sentences, i.e., negative and positive counterparts, with otherwise identical or similar meanings, from the domain of restaurant reviews. 500 sentences were originally written as positive and manually transferred to negative, the other 500 went in the opposite direction. The data is available in English and Bengali, with English originally based on (Li et al., 2018). However, the English data are not identical, as Mukherjee et al. (2023a) revised the texts to address issues like inconsistencies, spelling errors, inaccuracies in sentence sentiment, compromised linguistic fluency, omitted context, and improper sentiment adjustments.

We translated the English dataset into eight Indian languages to serve the aims of our experiment. In the following subsections, we briefly overview the TST task's language selection process in Section 3.1. We also explore the manual style-translation process and the challenges encountered in Section 3.2.

### 3.1 Language Selection

As discussed earlier, the eight Indian languages, namely Hindi, Magahi, Marathi, Malayalam, Punjabi, Odia, Telugu, and Urdu, are chosen for the sentiment transfer tasks. Malayalam and Telugu represent the Dravidian language family, while the rest of the languages belong to the Indo-Aryan languages. All of these languages are motivated by their substantial online user base, geographical dominance of the languages (see Table 6 in Appendix A for a short overview of these languages), increasing engagement in native language communication on social media,[2] and/or the usage statistics of language as content on the web.[3] This includes writing online reviews in these languages, making the base English sentiment dataset (Li et al., 2018) a suitable match for our study.

In addition, the choice of languages is also based on their affinities and differences in scripts, lexical and syntactic structure, and language families. All these, except Magahi, are among the 22 scheduled (official) Indian languages (Jha, 2010). Magahi, closely related to Hindi but distinct, presents an opportunity to explore multilingual sentiment transfer for a language with a limited internet presence. Odia and Hindi use different scripts but have common typological features and share lexical words due to belonging to the same language family (Ojha et al., 2015). Similarly, despite their close linguistic similarity, Urdu and Hindi exhibit notable differences in script and lexical composition. The linguistic diversity within this set of languages, including script variations and familial connections, can provide comparative analysis in style transfer from the linguistics perspective, including cultural nuances.

[2] https://assets.kpmg.com/content/dam/kpmg/in/pdf/2017/04/Indian-languages-Defining-Indias-Internet.pdf

[3] https://w3techs.com/technologies/overview/content_language

### 3.2 Style Translation Process

Qualified language experts or linguists working with a professional service provider for linguistic services were engaged for the translation (see Appendix A for the linguists' demographics and precise guidelines to maintain style accuracy and quality). Every language utilized a team comprising one translator and one validator, both native speakers.

The primary challenges we encountered in the process are described below, and more examples and their corresponding analyses are presented in Tables 13 and 14 in Appendix D. Some Sentiment transfer task-specific challenges are as follows:

**Implicit sentiment** Sentences where the sentiment is not expressed directly but as a result of an event or situation. For example, in the *my toddler found a dead mouse under one of the seats* sentence, sentiment is carried by the event of finding a dead mouse, hinting at the cleanliness and hygiene issues. Therefore, the context was removed and written as, *the place is clean and hygienic for kids and toddlers*.

**Insufficient context** Lack of context poses a problem in preserving the sentiment. For example, the phrase *sounds good doesn't it ?*, presented in isolation in the English dataset, looks like the tail end of another comment. Translating such sentences can lead to individual interpretations of context and sentiment variations.

**Fuzzy expressions** Although words like *um, uh* etc successfully lend positivity or negativity to a sentence, they leave a lot to one's imagination, further causing multiple interpretations. For example, in the sentence *i replied, "um... no i'm cool*, the expression *um* can be translated either as bad or ordinary or exciting.

**Suitable sentiment** There are instances when an English source sentence must be translated specifically to preserve the sentiment, not as a general translation. For example, the English sentence *no thanks amanda, i won't be back !* would be translated normally धन्यवाद अमांडा, मैं वापस नहीं आऊँगा! to Hindi, which is *thanks amanda, i won't be back!* in English. However, to preserve the negative sentiment style and content, the idiom भाड़ में जाओ is used in Hindi, which would map to *go to hell* in English.

**Confounding Phrase Structure** The data primarily concerns food, eating experience, and restaurants. Hence, there are a considerable number of dishes and their descriptions. The translation exercise has had difficulty decoding the dishes' names as either *adj+proper noun* or adjective as part of the proper noun phrase. For instance, if *[hot Thai basil soup]* could be *hot [thai basil] soup*, or *[hot] thai basil soup* and could be translated into Hindi like गर्म थाई–बेसिल सूप or गर्म थाई बेसिल सूप.

We also list some general translation-related challenges that we encountered:

**Gender encoding** Personal pronouns in English can be replaced with demonstrative pronouns in Indo-Aryan languages, thus removing gender information. On the contrary, certain verb phrases will have to take a gender role, which is otherwise missing in English. Thus, even when an English sentence did not encode any gender information, Indo-Aryan languages were forced to encode gender. For instance, in the sentence *just left and took it off the bill*, the gender is encoded in the verb, making it either masculine or feminine.

**Ambiguities** Ambiguity is a core feature of all languages and creates a challenge while translating, e.g., the word *cool* in the sentence *The environment here is cool* can be interpreted as either cold or filled with fun.

**Cultural references** Phrases like *corn people* can be challenging for translators who do not share American cultural references in their languages.

**Lexical gap** There are no direct translations of words like *pushy, welcoming, brunch, unwelcoming,* and *accommodating* in all target languages. Therefore, close approximations were chosen to maintain the sentiment.

**Noun anchoring** There are certain adjectives in English that work without the support of their nouns, e.g. *unfriendly and unwelcoming with a bad atmosphere and food*. In Indo-Aryan languages, noun support is mandatory and a linguistic equivalent of *behaviour* must be added.

| Challenges | Frequency (%) |
|---|---|
| Ambiguities | 34.0 |
| Lexical gap | 31.0 |
| Gender encoding | 30.0 |
| Cultural references | 21.0 |
| Insufficient context | 19.5 |
| Implicit sentiment | 19.0 |
| Lack of punctuation | 12.5 |
| Idiomatic expressions | 07.5 |
| Fuzzy expressions | 07.0 |
| Noun anchoring | 07.0 |
| Suitable sentiment | 06.0 |

Table 1: Statistics (approximate) of the challenges faced during datasets preparation, see details in Section 3.2.

**Lack of punctuation** Several texts join multiple independent phrases together with no punctuation, e.g., *i had a spanish omelet was huge and delicious.* The lack of punctuation makes it unnatural when translated into Indian languages.

**Idiomatic expressions** Phrases like *kicks ass*, or expressions like *sparkling wine flights* run the risk of being incorrectly translated if the translator is unaware of their idiomatic meanings, particularly the cultural context of the different countries/regions.

The approximate frequency of the aforementioned individual issues across all languages is illustrated in Table 1. Issues with *Ambiguities*, *Gender encoding*, and *Lexical gap* occurred most frequently.[4] For additional details, see Appendix E.

## 4 Models

Our experimental models use five methodologies (Sections 4.1-4.5): parallel, non-parallel, cross-lingual, shared multilingual learning and prompted LLMs. The first three methods are adopted from Mukherjee et al. (2023a), and we only briefly summarize them. The last two are newly introduced for this task.

### 4.1 Parallel Style Transfer

In this experiment (labeled *Parallel*), we fine-tune a pre-trained multilingual BART model (mBART) (Liu et al., 2020) using the parallel datasets constructed in Section 3.

[4]The distribution across target languages is roughly the same except for *Gender encoding*, which is highly-language dependent (in Odia, Malayalam, and Magahi, gender does not need to be coded).

### 4.2 Non-parallel Style Transfer

In this experiment, we focus on one part of the data at a time (positive/negative), building two separate models trained to produce sentences of a given sentiment. This approach leverages a scenario where parallel datasets are unavailable. We use four different model variants:

**Reconstruction through Auto-encoder and Back-translation** We use input reconstruction via an auto-encoder (*AE*) (Shen et al., 2017; Li et al., 2021) and back-translation (*BT*) (Prabhumoye et al., 2018b; Mukherjee et al., 2022). Each model is trained for a single sentiment. During inference, a sentence with the opposite sentiment is input to the model trained for the target sentiment (e.g., a positive sentence is input to the AE or BT model trained for negative sentence reconstruction). For BT, English sentences undergo an English-to-Hindi-to-English cycle, while other languages use source-to-English-to-source translation (for translations' experimental results, see Table 9 in Appendix C).

**Masked Style Filling (*MSF*)** By masking style-specific words in the input sentence, we enhance AE and BT with Masked Style Filling (*MSF-AE, MSF-BT*). Significant style-specific words are identified using integrated gradients (Sundararajan et al., 2017; Janizek et al., 2021) from our fine-tuned sentiment classification models (see Section 5.3). Words contributing most to sentiment are masked, making sentences "style-independent". These modified sentences are then used as input for *AE* and *BT* models to reconstruct the original sentences.

### 4.3 Cross-Lingual Style Transfer

We explore two cross-lingual alternatives that bypass the requirement for manually created multilingual datasets. Firstly, we employ English sentences from the parallel dataset, machine-translate them into all the respective languages, and use these translated texts for training (*En-IP-TR-Train*). Secondly, we take the English output generated by the model trained on a parallel English dataset and machine-translate it into the target languages (*En-OP-TR*). These cross-lingual approaches offer insights into multilingual text style transfer for the case when no data is available in the target languages.

### 4.4 Shared Learning Style Transfer

We conducted a joint training (*Joint*) following the *Parallel* approach (see Section 4.1), using style-parallel data from all the languages together. Despite the linguistic diversity, these languages have commonalities and shared characteristics. Learning them together enhances the availability of resources and helps exchange information across languages, benefiting the TST task overall. We introduced distinct language identifier prefixes and added them as special tokens for the model to treat them separately. For instance, for English, we used <*en*>, and for Hindi, we utilized <*hi*>, etc.

### 4.5 Large Language Models

For our experiments, we chose the *Llama2* and *Llama2_chat* models (Touvron et al., 2023a,b), each with 7B parameters and available under an open license on HuggingFace (Wolf et al., 2020). We also included *GPT-3.5* (*gpt-3.5-turbo–0125*) accessed via the OpenAI API (OpenAI, 2023). We used few-shot prompting for these models (for example, see Table 12 in Appendix C).

## 5 Experimental Details

### 5.1 Used Models & Language Support

For generating transferred text with the target style in all text-to-text generation processes in Section 4, we used *mBART-large-50* (Tang et al., 2020). We used *NLLB-200* (Costa-jussà et al., 2022) for the translation process involved in Sections 4.2 and 4.3. *XLM-RoBERTa-base* (Conneau et al., 2019) was used for multilingual sentiment classifications in Section 5.3. For evaluating embedding similarity, we used *LaBSE* (Feng et al., 2022), and for fluency calculation in terms of PPL in Section 6, we used *mGPT* (Shliazhko et al., 2024).[5]

Table 7 in Appendix C lists the supported languages for all models.

### 5.2 Settings

Each dataset comprises 1,000 style-parallel examples (see Section 3). To ensure consistency in our experiments, we divided these into 400 training examples, 100 for development, and 500 for testing.

Since parameter optimization for all languages model-wise would be resource-intensive and time-consuming, we optimized parameters for all languages only for the *Parallel* Methodology (see Section 4.1) and applied those settings to other methodologies for each language (in Appendix C).

For the MSF experiments (Section 4.2), we implemented a threshold of 0.25 to selectively filter style lexicons, determined via experiments on Hindi and English and applied to all languages (see Appendix C).

[5]All models were downloaded from HuggingFace (Wolf et al., 2020).

| Language | Sentiment Accuracy (%)↑ |
|---|---|
| English | 92.5 |
| Hindi | 89.9 |
| Magahi | 88.0 |
| Malayalam | 88.3 |
| Marathi | 90.0 |
| Odia | 84.3 |
| Punjabi | 87.9 |
| Telugu | 85.0 |
| Urdu | 87.4 |

Table 2: Language-wise sentiment classifier accuracy scores.

### 5.3 Multilingual Sentiment Classification

In our MSF experiments (see Section 4.2) and for evaluating sentiment transfer accuracy in all experiments (see Section 6), we fine-tuned an individual sentiment classifier for each language based on the *XLM-RoBERTa-base* model (Conneau et al., 2019), using the same training datasets as for our primary TST task (for results on batch optimization, see Table 8 in Appendix C). Table 2 presents the resulting classifier accuracies of individual languages.

## 6 Evaluation Metrics

The evaluation process comprises three critical dimensions: sentiment transfer accuracy, content retention, and linguistic fluency. We employed our fine-tuned classifiers to calculate *sentiment transfer accuracy (ACC)* (see Section 5.3). In line with previous studies (Mukherjee et al., 2023b,c; Jin et al., 2022; Hu et al., 2022), we evaluate *content preservation* through the BLEU score (Papineni et al., 2002) and *embedding similarity (CS)* (Rahutomo et al., 2012) when compared to the input sentences. The embedding similarity (CS) is computed using LaBSE sentence embeddings (Feng et al., 2022) in combination with cosine similarity. Similarly to Loakman et al. (2023) and Yang and Jin (2023), we derive a single comprehensive score for the two important measures of TST, *sentiment transfer accuracy* and *content preservation*, by calculating the arithmetic mean (AVG) (Mukherjee et al., 2022) of ACC, BLEU, and CS. While this is not ideal, as the scores' sensitivities are different, it allows us to easily compare with an accuracy-preservation tradeoff.

Assessing linguistic fluency, particularly for all the Indian languages, presents a challenge due to the absence of robust evaluation tools for Indian languages (Krishna et al., 2022). Earlier research cautioned against using perplexity (PPL) as a measure of fluency, as it tends to favor awkward sentences with commonly used words (Pang, 2019; Mir et al., 2019). With this in mind, we still present a basic fluency evaluation using PPL with a multilingual GPT (mGPT) model (Shliazhko et al., 2024).

All experiments were conducted separately for positive-to-negative and negative-to-positive sentiment transfer tasks. The metric results were then averaged and presented in this paper.

As automated metrics for language generation may not correlate well with human judgments (Novikova et al., 2017), we also run a small-scale human evaluation with expert annotators, i.e., the same linguists that were involved in the dataset creation process, on a random sample of 50 sentences from the test set for selected models (equally split to both positive-to-negative and negative-to-positive sentiment transfer tasks). The outputs are rated on a 5-point Likert scale for style transfer accuracy, content preservation, and fluency (for details, see Appendix B).

## 7 Results and Analysis

### 7.1 Automatic Evaluation

Table 3 presents automatic metric results for all languages. We describe the performance of the individual model types and contrast different languages.

**Parallel Style Transfer** The *Parallel* model, which leverages style-parallel datasets, shows balanced overall performance with strong scores on all three main metrics, indicating its effectiveness in preserving the content while changing its sentiment. These results highlight the benefits of using parallel datasets, even with a few training examples. While the accuracy stays relatively strong in most languages, it drops slightly for Punjabi and Odia. This difference may indicate that style transfer is more challenging in these languages or that the underlying multilingual pre-trained model has not been sufficiently exposed to them.

**Non-parallel Style Transfer** Non-parallel models generally perform worse than parallel ones. The Auto-Encoder (AE) model excels in content preservation but falls short of reaching the target style. Conversely, the Back-Translation (BT) model shows better style transfer accuracy but struggles with content preservation. This could be because back-translation tends to lose source stylistic attributes, which helps transfer them to the target style, but it may also lose original content, affecting content preservation (Mukherjee et al., 2022). The MSF extension improves results for both AE and BT models, enhancing style accuracy and fluency. However, it still struggles with BLEU scores, indicating challenges in content preservation.

**Cross-Lingual Style Transfer** Both models, *En-IP-TR-Train* (training on translated English data) and *En-OP-TR* (translating the English model's output), yield very competitive results in terms of style accuracy and content preservation. This showcases the potential of using machine translation of the style-parallel English data for TST tasks when an actual TST dataset is unavailable in the target language.

**Shared Learning Style Transformation** The *Joint* model, where all languages are trained together, exhibits strong performance in sentiment accuracy and content preservation. This is especially notable for English, Malayalam, Telugu, and Urdu, where this variant offers the best results, surpassing the language-specific *Parallel* model. These results highlight the benefits of shared learning in TST across multiple languages, suggesting that training in diverse languages can enhance model performance.

**Large Language Models** GPT-3.5 leads in overall performance. However, we can achieve comparable results with simpler, smaller, open models and minimal data. Our models deliver better-balanced results for Malayalam, Urdu, Magahi, Odia, and Telugu than GPT-3.5. This suggests that dedicated approaches and style-parallel data can sometimes outperform even LLMs, especially for low-resourced languages. Llama2 and Llama2_chat show average results in English and Hindi and poor results in all other languages.

**Language-wise Analysis** While the absolute scores in English and non-English languages are not directly comparable, overall, the comparatively lower values for sentiment transfer accuracy and content preservation in non-English languages (except Hindi) indicate that TST is more challenging for multilingual LMs in these languages. Variations in performance can be attributed to language-specific characteristics, data availability, and the extent to which pre-trained models have been trained with data from these languages. Hindi, as an exception among the non-English languages, performs relatively well due to its status as a resource-rich language (Joshi et al., 2020) with significant pretraining data available. This results in higher sentiment accuracy and content preservation than other non-English languages. In contrast, low-resource languages such as Marathi, Magahi, and Odia face more challenges. However, we note that lower BLEU for content preservation in these languages could be attributed to their complex linguistic properties and the strict nature of BLEU, which focuses on exact word overlap.

| Models | English | | | | | Hindi | | | | | Magahi | | | | |
|---|---|---|---|---|---|---|---|---|---|---|---|---|---|---|---|
| | ACC↑ | BLEU↑ | CS↑ | PPL↓ | AVG↑ | ACC↑ | BLEU↑ | CS↑ | PPL↓ | AVG↑ | ACC↑ | BLEU↑ | CS↑ | PPL↓ | AVG↑ |
| Parallel | 79.5 | 46.5 | 81.5 | 102.3 | 69.2 | 86.5 | 44.5 | 82.5 | 8.7 | 71.2 | 81.5 | 38.5 | 74.5 | 37.1 | 64.8 |
| AE | 7.5 | 42.0 | 78.0 | 102.3 | 42.5 | 10.0 | 41.5 | 80.0 | 8.9 | 43.8 | 12.0 | 36.5 | 71.5 | 37.3 | 40.0 |
| BT | 27.0 | 11.5 | 65.5 | 118.0 | 34.7 | 24.5 | 8.0 | 72.0 | 9.4 | 34.8 | 32.5 | 2.5 | 51.0 | 26.3 | 28.7 |
| MSF-AE | 64.5 | 36.0 | 72.5 | 200.2 | 57.7 | 65.5 | 29.0 | 72.0 | 9.0 | 55.5 | 80.5 | 25.0 | 63.0 | 38.1 | 56.2 |
| MSF-BT | 67.0 | 8.0 | 56.5 | 65.7 | 43.8 | 67.5 | 5.5 | 65.5 | 7.7 | 46.2 | 72.0 | 1.0 | 44.0 | 25.0 | 39.0 |
| En-IP-TR-Train | | | - | | | 79.0 | 41.0 | 81.5 | 8.7 | 67.2 | 69.5 | 31.0 | 71.0 | 31.7 | 57.2 |
| En-OP-TR | | | - | | | 78.5 | 14.0 | 77.0 | 8.0 | 56.5 | 77.5 | 4.5 | 59.5 | 21.7 | 47.2 |
| Joint | 86.5 | 42.0 | 81.0 | 56.2 | 69.8 | 76.0 | 43.5 | 79.0 | 24.6 | 66.2 | 87.0 | 31.0 | 75.5 | 19.7 | 64.5 |
| Llama2 | 25.0 | 43.0 | 78.5 | 114.2 | 48.8 | 50.0 | 34.0 | 74.5 | 9.9 | 52.8 | 31.5 | 32.0 | 66.0 | 37.7 | 43.2 |
| Llama2_chat | 88.0 | 37.0 | 77.5 | 87.7 | 67.5 | 56.5 | 34.5 | 73.0 | 9.3 | 54.7 | 36.0 | 31.5 | 63.5 | 33.4 | 43.7 |
| GPT-3.5 | 93.5 | 45.0 | 81.5 | 88.3 | 73.3 | 91.5 | 41.0 | 82.5 | 7.5 | 71.7 | 84.5 | 36.5 | 73.0 | 31.7 | 64.7 |

| Models | Malayalam | | | | | Marathi | | | | | Odia | | | | |
|---|---|---|---|---|---|---|---|---|---|---|---|---|---|---|---|
| | ACC↑ | BLEU↑ | CS↑ | PPL↓ | AVG↑ | ACC↑ | BLEU↑ | CS↑ | PPL↓ | AVG↑ | ACC↑ | BLEU↑ | CS↑ | PPL↓ | AVG↑ |
| Parallel | 78.5 | 25.0 | 77.0 | 4.9 | 60.2 | 79.5 | 26.0 | 78.5 | 8.6 | 61.3 | 63.0 | 28.0 | 76.5 | 2.2 | 55.8 |
| AE | 11.5 | 24.5 | 76.0 | 4.8 | 37.3 | 10.0 | 25.0 | 77.0 | 9.4 | 37.3 | 15.5 | 28.0 | 77.0 | 2.2 | 40.2 |
| BT | 30.0 | 3.5 | 64.5 | 6.2 | 32.7 | 28.5 | 5.0 | 66.5 | 10.9 | 33.3 | 86.5 | 2.0 | 48.0 | 2.2 | 45.5 |
| MSF-AE | 58.5 | 17.5 | 66.0 | 9.9 | 47.3 | 79.5 | 16.0 | 66.5 | 9.9 | 54.0 | 87.5 | 20.5 | 69.0 | 2.2 | 59.0 |
| MSF-BT | 72.0 | 2.0 | 59.5 | 5.6 | 44.5 | 73.0 | 3.5 | 59.5 | 9.4 | 45.3 | 96.0 | 1.5 | 47.0 | 2.0 | 48.2 |
| En-IP-TR-Train | 78.5 | 28.0 | 79.5 | 6.7 | 62.0 | 62.0 | 26.5 | 77.0 | 5.9 | 55.2 | 37.5 | 33.5 | 78.0 | 2.5 | 49.7 |
| En-OP-TR | 72.0 | 22.5 | 75.0 | 4.9 | 56.5 | 64.0 | 25.0 | 78.0 | 8.8 | 55.7 | 45.5 | 25.5 | 76.5 | 2.2 | 49.2 |
| Joint | 79.0 | 9.5 | 75.0 | 5.1 | 54.5 | 77.5 | 13.0 | 78.0 | 8.3 | 56.2 | 77.5 | 10.0 | 75.0 | 2.1 | 54.2 |
| Llama2 | 29.5 | 12.5 | 62.5 | 6.0 | 34.8 | 30.5 | 18.0 | 68.5 | 9.4 | 39.0 | 39.5 | 6.0 | 48.5 | 2.4 | 31.3 |
| Llama2_chat | 29.5 | 11.0 | 58.0 | 6.1 | 32.8 | 39.0 | 19.0 | 69.5 | 9.8 | 42.5 | 38.5 | 7.0 | 51.0 | 2.4 | 32.2 |
| GPT-3.5 | 75.0 | 23.5 | 75.5 | 4.8 | 58.0 | 83.0 | 24.5 | 79.0 | 9.4 | 62.2 | 76.5 | 23.5 | 72.5 | 2.2 | 57.5 |

| Models | Punjabi | | | | | Telugu | | | | | Urdu | | | | |
|---|---|---|---|---|---|---|---|---|---|---|---|---|---|---|---|
| | ACC↑ | BLEU↑ | CS↑ | PPL↓ | AVG↑ | ACC↑ | BLEU↑ | CS↑ | PPL↓ | AVG↑ | ACC↑ | BLEU↑ | CS↑ | PPL↓ | AVG↑ |
| Parallel | 63.0 | 36.0 | 78.5 | 2.6 | 59.2 | 70.5 | 23.5 | 72.5 | 6.2 | 55.5 | 71.5 | 34.0 | 79.5 | 31.5 | 61.7 |
| AE | 12.0 | 35.0 | 78.0 | 2.6 | 41.7 | 15.0 | 25.5 | 74.0 | 6.1 | 38.2 | 12.5 | 33.0 | 79.0 | 33.1 | 41.5 |
| BT | 78.0 | 5.0 | 55.5 | 14.0 | 46.2 | 33.5 | 3.0 | 63.5 | 7.6 | 33.3 | 24.5 | 8.5 | 69.5 | 71.5 | 34.2 |
| MSF-AE | 84.0 | 25.5 | 68.0 | 3.4 | 59.2 | 67.0 | 15.5 | 63.5 | 6.0 | 48.7 | 63.5 | 23.5 | 71.5 | 38.3 | 52.8 |
| MSF-BT | 95.5 | 3.0 | 48.5 | 2.5 | 49.0 | 62.0 | 2.5 | 59.0 | 5.9 | 41.2 | 73.0 | 6.0 | 63.5 | 84.2 | 47.5 |
| En-IP-TR-Train | 56.0 | 29.0 | 75.5 | 4.4 | 53.5 | 69.5 | 32.0 | 79.0 | 16.2 | 60.2 | 86.5 | 40.5 | 80.5 | 62.7 | 69.2 |
| En-OP-TR | 56.0 | 34.0 | 76.5 | 2.6 | 55.5 | 52.0 | 23.0 | 74.0 | 6.0 | 49.7 | 69.0 | 32.5 | 79.5 | 34.3 | 60.3 |
| Joint | 79.5 | 18.5 | 76.5 | 2.5 | 58.2 | 77.0 | 6.0 | 73.0 | 6.2 | 52.0 | 77.5 | 20.5 | 79.5 | 50.0 | 59.2 |
| Llama2 | 35.0 | 12.0 | 54.5 | 2.9 | 33.8 | 38.0 | 5.0 | 49.5 | 6.7 | 30.8 | 45.0 | 27.0 | 72.5 | 48.2 | 48.2 |
| Llama2_chat | 33.0 | 12.0 | 55.5 | 2.9 | 33.5 | 39.0 | 5.5 | 50.0 | 6.7 | 31.5 | 55.0 | 27.0 | 72.0 | 47.2 | 51.3 |
| GPT-3.5 | 85.5 | 34.5 | 78.5 | 2.6 | 66.2 | 70.5 | 23.0 | 74.5 | 5.9 | 56.0 | 87.0 | 32.5 | 80.5 | 31.7 | 66.7 |

Table 3: Automatic evaluation results. We measure the sentiment classifier accuracy (ACC), BLEU score, content similarity (CS), fluency (PPL), and the average (AVG) of ACC, BLEU, and CS (For details, see Section 6). We have several models (see Section 4): *Parallel* that uses parallel data, *AE* and *BT* for non-parallel data trained using input reconstruction, with extensions *MSF-AE* and *MSF-BT* employing masked style filling. *En-IP-TR-Train* trains on data machine-translated from English into the respective languages. *En-OP-TR* is machine translation of English model outputs. *Joint* refers to training a single multilingual model with all available data. Llama2, Llama2_chat and GPT-3.5 are off-the-shelf prompted LLMs. The best results in each category are highlighted in color.

| Models | English | | | Hindi | | | Magahi | | |
|---|---|---|---|---|---|---|---|---|---|
| | Style↑ | Content↑ | Fluency↑ | Style↑ | Content↑ | Fluency↑ | Style↑ | Content↑ | Fluency↑ |
| Parallel | 4.02 | 4.94 | 4.92 | 4.04 | 4.98 | 4.92 | 4.22 | 4.84 | 4.96 |
| Joint | 4.32 | 4.92 | 4.94 | 4.08 | 4.94 | 4.86 | 3.76 | 4.92 | 4.98 |
| GPT-3.5 | 4.56 | 4.98 | 4.96 | 4.68 | 4.98 | 4.90 | 3.96 | 4.90 | 4.62 |

Table 4: Human evaluation of 50 randomly selected outputs on style transfer accuracy (Style), Content Preservation (Content), and Fluency (see Section 6). The best results overall are highlighted in color.

While showing solid performance with certain models, Dravidian languages like Malayalam and Telugu still encounter difficulties, especially in maintaining BLEU scores. This suggests that structural differences in language families can influence the performance of sentiment transfer models. Despite achieving good results with specific models, these languages struggle with content preservation, indicating that their structure may pose more challenges for TST.

In conclusion, our experiments, particularly with the *Parallel* and *Joint* methodologies, underline the significance of parallel data in TST. The results of the MSF approach show that sentiment transfer accuracy can be improved in scenarios without parallel data, but performance remains worse than with parallel data. Cross-lingual models show that above-average results can be achieved without actual language-specific data, using high-quality MT from English. For additional details, see Appendix E.

### 7.2 Human Evaluation

For human evaluation, we selected our two best models: *Parallel* (see Section 4.1) and *Joint* (see Section 4.4), along with *GPT-3.5* (see Section 4.5), across three languages: English, Hindi, and Magahi, from Table 3 for their balanced performance on automatic metrics. The results, shown in Table 4, align closely with our automatic evaluation findings, validating the effectiveness of the data and experimented approaches. All models performed well in English across all metrics, with GPT-3.5 slightly leading in style and maintaining near-perfect scores in content preservation and fluency. In Hindi, GPT-3.5 excelled with the highest style score, but all models performed similarly in content preservation, and our Parallel model performed slightly better in fluency. For the low-resource language Magahi, the Parallel model achieved the highest style score, while our Joint model outperformed in content and fluency, surpassing GPT-3.5.

### 7.3 Generated Output Examples

Table 5 includes output samples for all the languages, using the same models as in Section 7.2, showing that sentiment transfer generally works well for most languages (English, Hindi, Magahi, Marathi, Telugu, and Urdu). The transfer is mostly accurate for Malayalam, although there are some instances where the nuance might slightly shift. Punjabi and Odia show inconsistencies. While the sentiment change is sometimes achieved, the context might be lost or altered significantly. Our Parallel and Joint models and GPT-3.5 show strong, comparable performance across multiple languages, often providing contextually and sentimentally accurate translations. Our Joint model outperforms GPT-3.5 in low-resource languages like Marathi and Punjabi. Additionally, our model's output closely matches human sentiment for Malayalam and Urdu, unlike GPT-3.5, which sometimes alters the intended meaning.

## 8 Conclusion

In this study, we address the problem of text style transfer, primarily focusing on multilingual TST in Indian languages. This work provides useful resources for TST in eight languages, explores various benchmark models, and presents an analysis of experimental results for all these languages. Furthermore, it is worth noting that our presented datasets are style-parallel and parallel across the languages, making them consistent and comparable for the TST task. In future work, we plan to explore a wider range of style attributes and incorporate more languages, leveraging our existing methodologies and framework, which can be adapted to any style attribute given the availability of parallel data.

| Model | Negative → Positive | Positive → Negative |
|---|---|---|
| Reference | first time i came in i knew i just wanted to leave. → first time i came in, i knew i just wanted something new.<br>hi: पहली बार जब मैं आया तो मुझे पता था कि मैं बस यहाँ से जाना चाहता था। → पहली बार जब मैं अंदर आया, तो मुझे पता था कि मुझे बस कुछ नया चाहिए।<br>mag: जब हम पहिला बार ऐली, तऽ हमरा पता हल कि हम बस निकलल चाहली। → पहिला बार हम अंदर ऐली, हमरा पता हल कि हम बस कुछ नया चाहित हि ।<br>mr: जेव्हा मी पहिल्यांदा आत आलो तेव्हा मला माहित होते की मला फक्त निघायचे आहे. → पहिल्यांदा मी आत आलो तेव्हा मला माहित होतं की मला काहीतरी नवीन हवं आहे.<br>ml: ആദ്യമായി ഞാൻ വന്നപ്പോൾ എനിക്ക് പോകണമെന്ന് അറിയാമായിരുന്നു. → ആദ്യമായി ഞാൻ വന്നപ്പോൾ, എനിക്ക് പുതിയ എന്തെങ്കിലും വേണമെന്ന് അറിയാമായിരുന്നു.<br>pa: ਪਹਿਲੀ ਵਾਰ ਜਦੋਂ ਮੈਂ ਅੰਦਰ ਆਇਆ ਤਾਂ ਮੈਨੂੰ ਪਤਾ ਸੀ ਕਿ ਮੈਂ ਬੱਸ ਛੱਡਣਾ ਚਾਹੁੰਦਾ ਸੀ। → ਪਹਿਲੀ ਵਾਰ ਜਦੋਂ ਮੈਂ ਅੰਦਰ ਆਇਆ, ਮੈਨੂੰ ਪਤਾ ਸੀ ਕਿ ਮੈਂ ਕੁਝ ਨਵਾਂ ਚਾਹੁੰਦਾ ਹਾਂ।<br>or: ପ୍ରଥମ ଥର ମୁଁ ଭିତରକୁ ଆସିଲି ମୁଁ ଜାଣିଥିଲି ଯେ ମୁଁ ଛାଡିବାକୁଚାହୁଁଛି। → ପ୍ରଥମ ଥର ମୁଁ ଭିତରକୁ ଆସିଲି, ମୁଁ ଜାଣିଲି ଯେ ମୁଁ କିଛିନୂଆ ଚାହୁଁଛି।<br>ur: میں پہلی بار آیا تھا مجھے معلوم تھا کہ میں صرف جانا چاہتا ہوں۔ → پہلی بار جب میں اندر آیا تھا، مجھے معلوم تھا کہ میں صرف کچھ نیا چاہتا ہوں۔<br>te: మొదటిసారి లోపలికి వచ్చాను, నాకు తెలుసు నేను వెళ్ళిపోవాలనుకుంటున్నాని. → మొదటిసారి నేను లోపలి వచ్చినప్పుడు నాకు తెలుసు నేనేదో కొంత కొత్తదనం కోరుకుంటున్నా. | thank you amanda, i will be back ! → no thanks amanda, i won't be back !<br>hi: धन्यवाद अमांडा, मैं वापस आऊंगा! → भाड़ में जाओ अमांडा, मैं वापस नहीं आऊँगा!<br>mag: धन्यवाद अमांडा, हम बापस आएम! → नऽ, धन्यवाद अमांडा, हम बापस नऽ आएम!<br>mr: धन्यवाद अमांडा, मी परत येईन! → नाही धन्यवाद अमांडा, मी परत येणार नाही!<br>ml: നന്ദി അമാൻഡ, ഞാൻ മടങ്ങിവരും! → ഇല്ല നന്ദി അമാൻഡ, ഞാൻ തിരികെ വരില്ല!<br>pa: ਧੰਨਵਾਦ ਅਮਾਂਡਾ ਵਾਪਸ ਆਵਾਂਗਾ! → ਕੋਈ ਧੰਨਵਾਦ ਨਹੀਂ ਅਮਾਂਡਾ, ਮੈਂ ਵਾਪਸ ਨਹੀਂ ਆਵਾਂਗਾ!<br>or: ଧନ୍ୟବାଦ ଅମାଣ୍ଡା, ମୁଁ ଫେରି ଆସିବି! → ନା ଧନ୍ୟବାଦ ଅମାଣ୍ଡା, ମୁଁ ଫେରିବି ନାହିଁ!<br>ur: آپ کا شکریہ امینڈا، میں واپس آؤں گا! → نہیں شکریہ امانڈا، میں واپس نہیں آؤں گا!<br>te: ధన్యవాదాలు అమండ, నేను వెనక్కి వస్తాను. → ధన్యవాదాలు కాదు అమండ, నేను వెనక్కి రాను. |
| Parallel | en: first time i came in I felt i felt right at home.<br>hi: पहली बार जब मैं आया तो मुझे पता था कि मैं बस यहाँ से जाना चाहता था।<br>mag: जब हम पहिला बार ऐली, तऽ हमरा पता हल कि हम बस निकलल चाहली।<br>mr: जेव्हा मी पहिल्यांदा आत आलो तेव्हा मला माहित होते की मी खूप चांगले काम करायचे आहे.<br>ml: ആദ്യമായി ഞാൻ വന്നപ്പോൾ എനിക്ക് പോകണമെന്ന് അറിയാമായിരുന്നു.<br>pa: ਪਹਿਲੀ ਵਾਰ ਜਦੋਂ ਮੈਂ ਅੰਦਰ ਆਇਆ ਤਾਂ ਮੈਨੂੰ ਪਤਾ ਸੀ ਕਿ ਮੈਂ ਬੱਸ ਛੱਡਣਾ ਚਾਹੁੰਦਾ ਸੀ।<br>or: ପ୍ରଥମ ଥର ମୁଁ ଭିତରକୁ ଆସିଲି ମୁଁ ଜାଣିଥିଲି ଯେ ମୁଁ ଛାଡିବାକୁଚାହୁଁଛି।<br>ur: میں پہلی بار آیا تھا مجھے معلوم تھا کہ میں صرف جانا چاہتا ہوں۔<br>te: మొదటిసారి లోపలికి వచ్చాను. | en: thank you amanda, i will not be back!<br>hi धन्यवाद अमांडा, मैं वापस नहीं जाऊंगा!<br>mag: धन्यवाद अमांडा, हम बापस आएम!<br>mr: धन्यवाद अमांडा, मी परत येणार नाही.<br>ml: അമാൻഡ, ഞാൻ തിരികെ വരില്ല!<br>pa: ਧੰਨਵਾਦ ਅਮਾਂਡਾ ਵਾਪਸ ਨਹੀਂ ਆਵਾਂਗਾ!<br>or: ଧନ୍ୟବାଦ ଅମାଣ୍ଡା, ମୁଁ ଫେରି ଆସିବି ନାହିଁ।<br>ur: آپ کا شکریہ امینڈا، میں واپس نہیں آؤں گا۔<br>te: ధన్యవాదాలు అమండ, నేను వెనక్కి వస్తాను. |
| Joint | en: first time I came in I knew I wanted to stay.<br>hi पहली बार जब मैं आया तो मुझे पता था कि मैं बस यहाँ से जाना चाहता था।<br>mag: जब हम पहिला बार ऐली, तऽ हमरा पता हल कि हम बस निकलल चाहली।<br>mr: जेव्हा मी पहिल्यांदा आत आलो तेव्हा मला माहित होते की मला फक्त निघायचे आहे.<br>ml: ആദ്യമായി ഞാൻ വന്നപ്പോൾ എനിക്ക് പോകണമെന്ന് അറിയാമായിരുന്നു.<br>pa: ਪਹਿਲੀ ਵਾਰ ਜਦੋਂ ਮੈਂ ਅੰਦਰ ਆਇਆ ਤਾਂ ਮੈਨੂੰ ਪਤਾ ਸੀ ਕਿ ਮੈਂ ਬੱਸ ਛੱਡਣਾ ਚਾਹੁੰਦਾ ਸੀ।<br>or: ପ୍ରଥମ ଥର ମୁଁ ଭିତରକୁ ଆସିଲି ମୁଁ ଜାଣିଥିଲି ଯେ ମୁଁ ଛାଡିବାକୁଚାହୁଁଛି।<br>ur: میں پہلی بار آیا تھا میں جانتا تھا کہ میں صرف جانا چاہتا ہوں۔<br>te: మొదటిసారి లోపలికి వచ్చాను, నాకు తెలుసు నేను వెళ్ళిపోవాలనుకుంటున్నాని. | en: sorry amanda, i will not be back.<br>hi धन्यवाद अमांडा, मैं वापस नहीं आऊंगा!<br>mag: धन्यवाद अमांडा, हम बापस नऽ आएम!<br>mr: धन्यवाद अमांडा, मी परत येणार नाही!<br>ml: നന്ദി അമാൻഡ, ഞാൻ തിരികെ വരില്ല!<br>pa: ਧੰਨਵਾਦ ਅਮਾਂਡਾ ਵਾਪਸ ਆਵਾਂਗਾ!<br>or: ଧନ୍ୟବାଦ ଅମାଣ୍ଡା, ମୁଁ ଫେରି ଆସିବି!<br>ur: آپ کا شکریہ نہیں، میں واپس نہیں آؤں گا!<br>te: ధన్యవాదాలు అమండ, నేను వెనక్కి రాను. |
| GPT3.5 | en: first time i came in i knew i never wanted to leave.<br>hi पहली बार जब मैं आया तो मुझे पता था कि मैं यहाँ रुकना चाहता था।<br>mag: जब हम पहिला बार ऐली, तऽ हमरा पता हल कि हम बस निकलल चाहली।<br>mr: जेव्हा मी पहिल्यांदा आत आलो तेव्हा मला माहित होते की मला आवडतंय आहे.<br>ml: ആദ്യമായി ഞാൻ വന്നപ്പോൾ എനിക്ക് പോകണമെന്ന് അറിയാമായിരുന്നു.<br>pa: ਪਹਿਲੀ ਵਾਰ ਜਦੋਂ ਮੈਂ ਅੰਦਰ ਆਇਆ ਤਾਂ ਮੈਨੂੰ ਪਤਾ ਸੀ ਕਿ ਮੈਂ ਬੱਸ ਆਗੇ ਬੜ੍ਹਣਾ ਚਾਹੁੰਦਾ ਸੀ।<br>or: ପ୍ରଥମ ଥର ମୁଁ ଭିତରକୁ ଆସିଲି ମୁଁ ଜାଣିଥିଲି ଯେ ମୁଁ ଛାଡିବାକ<br>ur: میں پہلی بار آیا تھا اور مجھے معلوم ہوا کہ میں اس جگہ کو بہت پسند کروں گا۔<br>te: మొదటిసారి లోపలికి వచ్చాను, నాకు తెలుసు నేను వెళ్ళిపోవాలనుకుంటున్నాని. | en: thank you amanda, i will never be back !<br>hi धन्यवाद अमांडा, मैं वापस नहीं आऊंगा।<br>mag: धन्यवाद अमांडा, हम बापस नऽ आएम।<br>mr: धन्यवाद अमांडा, मी परत येईन!<br>ml: നന്ദി അമാൻഡ, ഞാൻ മടങ്ങിവരുന്നില്ല!<br>pa: ਖੇਦ ਅਮਾਂਡਾ ਵਾਪਸ ਆਵਾਂਗਾ!<br>or: ଧନ୍ୟବାଦ ଅମାଣ୍ଡା, ମୁଁ ଫେରି ଆସିବି ନାହିଁ।<br>ur: آپ کا شکریہ امینڈا، میں واپس نہیں آؤں گا!<br>te: ధన్యవాదాలు అమండ, నేను వెనక్కి రాను. |

Table 5: Sample dataset references (on top, shown as input → target) and corresponding outputs generated by selected models for all languages in the dataset.

## Limitations

**Data Bias:** Our study relies on publicly available text data, which may inherently contain biases present in the sources from which it was collected. These biases can affect the performance of models trained on such data and may lead to biased outputs in sentiment transfer tasks.

**Generalization:** While our models perform well on our datasets, their ability to generalize to other domains or contexts may be limited.

**Subjectivity and Context:** Sentiment analysis is inherently subjective, and the sentiment labels assigned to sentences may not universally apply. The context in which a sentence is used can significantly influence its sentiment, and our models may not always capture nuanced contextual variations.

**Evaluation Metrics:** While we have employed a variety of evaluation metrics, including style transfer accuracy, content preservation, and fluency, no single metric captures all aspects of sentiment transfer. The evaluation process remains an active area of research, and further advancements in metrics may be needed.

## Ethics Statement

**Data Privacy and Consent:** We are committed to respecting data privacy and ensuring that all data used in our research is anonymized and devoid of personally identifiable information. We have taken measures to protect the privacy and confidentiality of individuals whose data may be included in our datasets.

**Bias Mitigation:** We acknowledge the potential presence of bias in our data sources and have taken steps to minimize the impact of such bias during model training and evaluation. We prioritize fairness and strive to mitigate any potential bias in our results.

**Transparency and Reproducibility:** We are dedicated to providing transparency in our research methods, including dataset collection, preprocessing, and model training. We encourage reproducibility by making our code and datasets publicly available.

**Informed Consent:** In cases where our research involves human annotators or data contributors, we have sought informed consent and followed ethical data collection and usage guidelines.

**Social Impact:** We recognize the potential social impact of our research and remain vigilant about the responsible use of AI technologies. We aim to contribute positively to the field of sentiment analysis and ensure our work benefits society as a whole.

## Acknowledgments

This research was funded by the European Union (ERC, NG-NLG, 101039303). We acknowledge the use of resources provided by the LINDAT/CLARIAH-CZ Research Infrastructure (Czech Ministry of Education, Youth, and Sports project No. LM2018101). We also acknowledge Panlingua Language Processing LLP for providing the dataset and this collaborative research project. Atul Kr. Ojha and John P. McCrae would like to acknowledge the support of the Science Foundation Ireland (SFI) as part of Grant Number SFI/12/RC/2289_P2 Insight_2, Insight SFI Research Centre for Data Analytics.

# A Data Statement

This section briefly provides the overview of the languages, translation guidelines, and demographics used to build the dataset (see Table 6, Section A.1 and A.2).

## A.1 Precise and General Guidelines

- As a language expert, you must translate the data into your language by following the consistency.
- This means you must translate both versions of each sentence.
- While translating, you must remember two primary principles:
  - One is that the translation should sound natural. The selection of words and phrases should be a natural way of speaking in your language.
  - Second is to preserve the maximum lexical, sentiment, and cultural context possible.
  - Wherever the principles come into conflict with each other, choose the first one.
- The sentences in the dataset include words that denote emotion or feelings that make the sentence either positive or negative. Do not skip those in your translation. For example, if “What the hell are you doing?” is translated as “Tum kya kar rahi ho?” the emotion is lost.

| Language | Language Family | Script | Regions | Speakers (in millions) |
|---|---|---|---|---|
| Hindi (hi) | Indo-Aryan | Devanagari | Uttar Pradesh, Bihar, Madhya Pradesh, Rajasthan, Haryana, Chhattisgarh, Jharkhand, Uttarakhand, West Bengal, Himachal Pradesh, Delhi, and Chandigarh | 528 |
| Magahi (mag) | Indo-Aryan | Devanagari | Bihar and some areas of Jharkhand, Odisha, and West Bengal | 12.6 |
| Malayalam (ml) | Dravidian | Brahami | Kerala, Lakshadweep and Puducherry | 34.8 |
| Marathi (mr) | Indo-Aryan | Devanagari | Maharashtra and Goa | 83 |
| Punjabi (pa) | Indo-Aryan | Gurumukhi | Punjab, Haryana and some areas of Jammu and Kashmir | 31.1 |
| Odia (or) | Indo-Aryan | Kalinga | Odisha and some areas Jharkhand and Bihar | 37 |
| Telugu (te) | Dravidian | Brahami | Andhra Pradesh, Telangana, Puducherry | 81.1 |
| Urdu (ur) | Indo-Aryan | Nastaliq | Uttar Pradesh, Bihar, Andhra Pradesh and Karnataka | 50 |

Table 6: Overview of the languages used in our experiment. We gathered speaker and spoken state statistics in Indian regions from the 2011 Census Report of India (https://censusindia.gov.in/nada/index.php/catalog/42458).

The word "hell" makes the sentence negative and should be included in the translated sentence.

- Use the comments section to write any challenges you face while translating a sentence, any heads up you want to provide to the reviewer, or anything incorrect was noticed.
- In certain situations, naturalness may demand transliteration of the English words. For example, blue cheese should be transliterated and not translated as 'neela cheese' in Hindi.

### A.2 Translators Demographic

- Hindi and English translator: with an M.Phil in Linguistics and an MA in English, native Hindi speaker and fluent in English, from Delhi, India.
- Magahi translator: with a PhD in Linguistics and native Magahi speaker and fluent in Hindi and English, from Bihar, India.
- Malayalam translator: with an MA in Linguistics and native Malayalam speaker and fluent in English, from Trivandrum, Kerala, India.
- Marathi translator: with an MA in Linguistics and native Marathi native speaker fluent in Hindi and English, from Mumbai, Maharashtra, India.
- Odia translator: with an MA in Linguistics and native Odia speaker, fluent in Hindi and English, from Bhubaneswar, Odisha, India.
- Punjabi translator: with an MA in Punjabi and native Punjabi speaker, fluent in Hindi and English, from Chandigarh, Punjab, India.
- Telugu translator: with MA in English and native Telugu speaker, fluent in Hindi and English, from Kuppam, Andhra Pradesh, India.
- Urdu translator: with MA in Urdu and native Urdu speaker, fluent in Hindi and English, from Sultanpur, Uttar Pradesh, India.

## B Human Evaluation Procedure

To evaluate the performance of our Text Sentiment Transfer models, we conducted a human evaluation focused on three critical aspects: *Style Transfer Accuracy*, *Content Preservation*, and *Fluency*. Below, we provide detailed definitions for each aspect and describe the questions used to guide the evaluation.

### B.1 Style Transfer Accuracy

**Definition:** Style Transfer Accuracy refers to how accurately the style of the original sentence has been transformed into the target sentiment. For instance, if a sentence originally expresses a negative sentiment, this metric evaluates whether it has been accurately converted to a positive sentiment, and vice versa.

**Evaluation Question:**

- *How accurately has the sentiment of the original sentence been transferred to the target sentiment?*

**Scoring:**

- **1**: No sentiment change; the original sentiment is entirely preserved.
- **2**: Minimal sentiment change; only slight evidence of sentiment transfer.
- **3**: Partial sentiment change; some aspects of the target sentiment are present, but the original sentiment still dominates.
- **4**: Considerable sentiment change; the target sentiment is clearly present, though traces of the original sentiment may remain.
- **5**: Complete sentiment change; the original sentiment has been entirely replaced by the target sentiment.

### B.2 Content Preservation

**Definition:** Content Preservation measures how well the style-independent meaning and core information of the original sentence are preserved after sentiment transfer.

**Evaluation Question:**

- *To what extent has the style-independent content and meaning of the original sentence been preserved after the sentiment transfer?*

**Scoring:**

- **1**: Content is completely altered; the original meaning is lost.
- **2**: Major content changes; significant parts of the original meaning are altered or missing.
- **3**: Moderate content preservation; the general idea is retained, but with some noticeable changes.
- **4**: Good content preservation; most of the original meaning is intact with only minor alterations.
- **5**: Complete content preservation; the original meaning is fully retained.

### B.3 Fluency

**Definition:** Fluency assesses the grammatical correctness, naturalness, and overall readability of the sentence after the sentiment transfer. A fluent sentence should flow naturally and be free of awkward constructions or errors.

**Evaluation Question:**

- *How fluent and natural does the sentence sound after the sentiment transfer?*

**Scoring:**

- **1**: Not fluent at all; the sentence is grammatically incorrect and difficult to understand.
- **2**: Limited fluency; the sentence contains multiple errors and reads awkwardly.
- **3**: Moderate fluency; the sentence is somewhat understandable but has noticeable issues.
- **4**: Good fluency; the sentence is mostly clear with only minor issues.
- **5**: Complete fluency; the sentence is grammatically correct, natural, and easy to read.

### B.4 Evaluation Process

Evaluators are asked to rate each of these aspects on a 5-point Likert scale for a random sample of 50 sentences from the test set, equally split between positive-to-negative and negative-to-positive sentiment transfer tasks.

## C Experimental Details

**Hyperparameter optimization:** To optimize the main generation mBART model's performance, we conducted hyperparameter tuning, selecting a learning rate 1e-5 and a separate batch size for each language experiment (see Table 10). Dropout was applied across the network at a rate of 0.1, and we introduced L2 regularization with a strength of 0.01. We trained the models for 30 epochs.

The MSF style-specific word selection threshold was chosen after experimenting with various values (see Table 11), and we found that using 0.25 resulted in a better balance between style transfer accuracy and content preservation in the target output.

## D Dataset and Generated Output Samples

In this section, we present a selection of samples from our curated datasets (see Table 14 and 13) along with generated output samples from selected models (see Table 5).

| Languages | Pre-trained models | | | | |
|---|---|---|---|---|---|
| | NLLB-200 | mBART-large-50 | BERT-base multilingual cased | LaBSE | mGPT |
| English | ✓ | ✓ | ✓ | ✓ | ✓ |
| Hindi | ✓ | ✓ | ✓ | ✓ | ✓ |
| Magahi | ✓ | ✗ | ✗ | ✗ | ✗ |
| Malayalam | ✓ | ✓ | ✓ | ✓ | ✓ |
| Marathi | ✓ | ✓ | ✓ | ✓ | ✓ |
| Odia | ✓ | ✗ | ✗ | ✓ | ✗ |
| Punjabi | ✓ | ✗ | ✓ | ✓ | ✗ |
| Telugu | ✓ | ✓ | ✓ | ✓ | ✓ |
| Urdu | ✓ | ✓ | ✗ | ✓ | ✓ |

Table 7: Languages covered by the pre-trained models used in this work. Some languages are not supported by some models, but they mostly share significant vocabulary and linguistic similarities with supported languages such as Hindi and others (Rudra et al., 2016; Kumar et al., 2018, 2021; Goswami et al., 2023; San et al., 2024).

| Batch size | English↑ | Hindi↑ | Magahi↑ | Malayalam↑ | Marathi↑ | Odia↑ | Punjabi↑ | Telugu↑ | Urdu↑ |
|---|---|---|---|---|---|---|---|---|---|
| 1 | 94.5 | 50.0 | 86.5 | 89.0 | 87.5 | 89.0 | 87.5 | 64.5 | 89.5 |
| 2 | 92.5 | 77.5 | 85.5 | 84.5 | 79.5 | 50.0 | 88.0 | 82.0 | 91.0 |
| 3 | 92.0 | 82.5 | 75.0 | 85.5 | 82.0 | 60.5 | 70.5 | 81.5 | 91.5 |
| 4 | 87.0 | 83.0 | 85.0 | 84.5 | 85.0 | 79.0 | 88.5 | 84.0 | 86.5 |
| 8 | 93.0 | 85.0 | 82.0 | 84.0 | 85.5 | 82.5 | 82.5 | 85.5 | 91.5 |
| 16 | 92.0 | 86.5 | 84.5 | 89.0 | 89.0 | 88.0 | 87.5 | 83.5 | 88.0 |
| 32 | 94.0 | 83.5 | 85.0 | 88.0 | 89.0 | 84.5 | 83.5 | 83.0 | 90.0 |
| 64 | 93.0 | 85.5 | 87.0 | 88.0 | 92.0 | 86.0 | 85.0 | 87.0 | 88.5 |

Table 8: Optimized batch-size finding results of the multilingual sentiment classifiers (see Section 5.3).

| MSF-BT | | | | En-IP-TR-Train | | En-OP-TR | |
|---|---|---|---|---|---|---|---|
| Task | BLEU↑ | Task | BLEU↑ | Task | BLEU↑ | Task | BLEU↑ |
| en→hi | 20.7 | en→hi→en | 42.6 | en→hi | 20.7 | en→hi | 17.1 |
| hi→en | 26.1 | hi→en→hi | 29.9 | en→mag | 06.4 | en→mag | 05.6 |
| mag→en | 18.1 | mag→en→mag | 07.9 | en→ml | 18.8 | en→ml | 12.1 |
| ml→en | 32.9 | ml→en→ml | 20.7 | en→mr | 25.9 | en→mr | 16.2 |
| mr→en | 32.4 | mr→en→mr | 27.3 | en→or | 18.3 | en→or | 12.4 |
| or→en | 33.1 | or→en→or | 21.8 | en→pa | 34.1 | en→pa | 23.8 |
| pa→en | 34.6 | pa→en→pa | 38.2 | en→te | 09.5 | en→te | 07.0 |
| te→en | 24.7 | te→en→te | 14.2 | en→ur | 38.9 | en→ur | 26.7 |
| ur→en | 38.4 | ur→en→ur | 40.9 | - | | - | |

Table 9: BLEU scores for *translations* used in Section 4.2 and 4.3.

# E Additional Dataset and Results Statistics

In this section, we present various graphs and charts derived from our datasets (see Section 3) and automatic evaluation results shown in Table 3 (and related to the analysis discussed in Section 7) to provide further insights.

| | English | | | | | Hindi | | | | | Magahi | | | | |
|---|---|---|---|---|---|---|---|---|---|---|---|---|---|---|---|
| **Batch** | ACC↑ | CS↑ | BLEU↑ | PPL↓ | AVG↑ | ACC↑ | CS↑ | BLEU↑ | PPL↓ | AVG↑ | ACC↑ | CS↑ | BLEU↑ | PPL↓ | AVG↑ |
| 1 | 75.5 | 79.5 | 43.0 | 116.9 | 66.0 | 79.5 | 81.5 | 43.5 | 10.2 | 68.2 | 76.5 | 71.5 | 37.0 | 44.5 | 61.7 |
| 2 | 73.0 | 79.0 | 43.0 | 159.6 | 65.0 | 88.0 | 81.5 | 43.0 | 10.4 | 70.8 | 80.5 | 71.0 | 35.0 | 45.0 | 62.2 |
| 3 | 81.5 | 79.5 | 43.0 | 120.2 | 68.0 | 88.5 | 81.5 | 43.5 | 10.7 | 71.2 | 82.0 | 72.0 | 36.5 | 43.8 | 63.5 |
| 4 | 79.0 | 79.5 | 42.5 | 106.3 | 67.0 | 74.5 | 80.5 | 43.5 | 10.6 | 66.2 | 75.0 | 72.0 | 36.0 | 44.7 | 61.0 |
| 8 | 75.0 | 78.5 | 41.5 | 112.5 | 65.0 | 79.5 | 82.0 | 44.5 | 10.3 | 68.7 | 80.0 | 70.5 | 35.0 | 42.3 | 61.8 |
| 16 | 71.0 | 78.5 | 41.0 | 124.1 | 63.5 | 78.5 | 81.5 | 44.0 | 10.3 | 68.0 | 76.5 | 71.0 | 37.0 | 44.8 | 61.5 |
| 32 | 65.0 | 69.0 | 25.5 | 668.4 | 53.2 | 83.5 | 81.5 | 43.0 | 9.9 | 69.3 | 81.5 | 71.5 | 36.5 | 42.5 | 63.2 |
| 64 | 66.5 | 56.0 | 10.0 | 275.2 | 44.2 | 81.0 | 82.5 | 45.5 | 10.3 | 69.7 | 74.5 | 72.0 | 36.5 | 43.6 | 61.0 |

| | Malayalam | | | | | Marathi | | | | | Odia | | | | |
|---|---|---|---|---|---|---|---|---|---|---|---|---|---|---|---|
| **Batch** | ACC↑ | CS↑ | BLEU↑ | PPL↓ | AVG↑ | ACC↑ | CS↑ | BLEU↑ | PPL↓ | AVG↑ | ACC↑ | CS↑ | BLEU↑ | PPL↓ | AVG↑ |
| 1 | 59.5 | 76.5 | 23.0 | 5.0 | 53.0 | 76.5 | 78.5 | 22.0 | 9.2 | 59.0 | 58.0 | 76.5 | 30.5 | 2.2 | 55.0 |
| 2 | 70.5 | 76.5 | 22.0 | 5.1 | 56.3 | 64.5 | 78.0 | 20.5 | 9.1 | 54.3 | 53.5 | 77.0 | 31.5 | 2.2 | 54.0 |
| 3 | 79.5 | 76.5 | 22.0 | 5.2 | 59.3 | 72.5 | 79.0 | 22.0 | 9.1 | 57.8 | 58.0 | 77.5 | 31.5 | 2.1 | 55.7 |
| 4 | 64.0 | 77.0 | 24.0 | 4.9 | 55.0 | 69.5 | 77.0 | 19.0 | 10.6 | 55.2 | 59.0 | 76.0 | 29.0 | 2.2 | 54.7 |
| 8 | 63.0 | 76.5 | 23.5 | 4.9 | 54.3 | 64.0 | 78.0 | 21.5 | 10.1 | 54.5 | 50.0 | 75.5 | 30.5 | 2.2 | 52.0 |
| 16 | 55.5 | 76.0 | 22.0 | 4.8 | 51.2 | 79.0 | 78.0 | 20.5 | 8.8 | 59.2 | 39.5 | 72.0 | 26.5 | 2.4 | 46.0 |
| 32 | 51.0 | 76.0 | 23.5 | 5.0 | 50.2 | 67.5 | 78.5 | 21.0 | 9.0 | 55.7 | 18.0 | 76.5 | 30.0 | 2.2 | 41.5 |
| 64 | 39.5 | 70.5 | 13.0 | 5.0 | 41.0 | 63.0 | 73.0 | 14.5 | 8.9 | 50.2 | 15.0 | 76.5 | 30.0 | 2.2 | 40.5 |

| | Punjabi | | | | | Telugu | | | | | Urdu | | | | |
|---|---|---|---|---|---|---|---|---|---|---|---|---|---|---|---|
| **Batch** | ACC↑ | CS↑ | BLEU↑ | PPL↓ | AVG↑ | ACC↑ | CS↑ | BLEU↑ | PPL↓ | AVG↑ | ACC↑ | CS↑ | BLEU↑ | PPL↓ | AVG↑ |
| 1 | 52.0 | 76.5 | 38.0 | 2.6 | 55.5 | 50.0 | 74.5 | 24.5 | 5.9 | 49.7 | 67.0 | 78.0 | 31.5 | 32.5 | 58.8 |
| 2 | 60.5 | 77.0 | 37.5 | 2.6 | 58.3 | 62.0 | 74.5 | 25.0 | 5.8 | 53.8 | 63.5 | 78.5 | 32.5 | 35.9 | 58.2 |
| 3 | 61.0 | 77.5 | 39.0 | 2.6 | 59.2 | 67.0 | 73.0 | 23.5 | 6.1 | 54.5 | 75.5 | 79.0 | 32.0 | 35.2 | 62.2 |
| 4 | 50.5 | 76.5 | 37.5 | 2.6 | 54.8 | 61.5 | 75.0 | 24.5 | 5.8 | 53.7 | 58.5 | 78.5 | 32.5 | 29.9 | 56.5 |
| 8 | 49.5 | 76.5 | 37.5 | 2.7 | 54.5 | 52.0 | 74.5 | 23.0 | 5.9 | 49.8 | 56.0 | 79.0 | 32.5 | 34.7 | 55.8 |
| 16 | 42.5 | 74.5 | 34.5 | 2.8 | 50.5 | 52.0 | 75.0 | 25.0 | 5.8 | 50.7 | 68.0 | 78.5 | 32.0 | 30.0 | 59.5 |
| 32 | 22.0 | 76.0 | 37.0 | 2.6 | 45.0 | 52.5 | 75.5 | 25.5 | 5.9 | 51.2 | 64.5 | 79.0 | 32.0 | 30.3 | 58.5 |
| 64 | 15.0 | 76.0 | 36.5 | 2.6 | 42.5 | 40.5 | 69.5 | 19.0 | 5.7 | 43.0 | 52.0 | 77.0 | 31.5 | 31.8 | 53.5 |

Table 10: Optimized batch-size finding results for each language using the *Parallel* (Section 4.1) methodology, for details see Section 5.2.

| | English | | | | | Hindi | | | | |
|---|---|---|---|---|---|---|---|---|---|---|
| **threshold** | ACC↑ | CS↑ | BLEU↑ | PPL↓ | AVG↑ | ACC↑ | CS↑ | BLEU↑ | PPL↓ | AVG↑ |
| *ae_mask* | | | | | | | | | | |
| 0.25 | 64.5 | 71.5 | 34.0 | 143.1 | 56.7 | 64.5 | 70.0 | 27.5 | 10.0 | 54.0 |
| 0.35 | 58.5 | 73.5 | 36.5 | 138.5 | 56.2 | 56.0 | 73.5 | 31.5 | 10.4 | 53.7 |
| 0.50 | 41.5 | 75.0 | 36.5 | 172.1 | 51.0 | 44.0 | 76.0 | 37.0 | 10.9 | 52.3 |
| 0.65 | 34.5 | 75.5 | 38.0 | 134.3 | 49.3 | 32.0 | 77.5 | 39.0 | 10.6 | 49.5 |
| 0.75 | 24.0 | 75.0 | 38.5 | 149.9 | 45.8 | 23.5 | 78.0 | 40.0 | 10.9 | 47.2 |
| *be_mask* | | | | | | | | | | |
| 0.25 | 69.5 | 56.0 | 7.5 | 72.0 | 44.3 | 68.0 | 64.5 | 4.5 | 8.6 | 45.7 |
| 0.35 | 56.5 | 56.5 | 8.5 | 92.1 | 40.5 | 64.5 | 66.0 | 5.5 | 8.1 | 45.3 |
| 0.50 | 37.5 | 61.5 | 9.5 | 92.8 | 36.2 | 47.0 | 67.5 | 5.5 | 8.0 | 40.0 |
| 0.65 | 43.0 | 62.5 | 11.0 | 105.2 | 38.8 | 46.5 | 67.5 | 7.0 | 9.5 | 40.3 |
| 0.75 | 35.0 | 62.5 | 11.0 | 106.9 | 36.2 | 37.5 | 67.5 | 7.0 | 9.9 | 37.3 |

Table 11: Optimized threshold finding results for selectively filtering style lexicons in MSF experiments (Section 4.2), for details see Section 5.2.

| **Prompt** | Sentiment transfer changes the sentiment of a sentence while keeping the rest of the content unchanged.<br>Examples:<br><br>Task: positive to negative<br>Input: जब उसने एकदम से कोई जवाब नहीं दिया, तो वह इत्मिनान से फ़ोन पर बना रहा ।<br>Output: जब उसने एकदम से कोई जवाब नहीं दिया, तो उसने फ़ोन काट दिया।<br><br>Task: negative to positive<br>Input: डेली में सलाद या पास्ता का अच्छा सिलेक्शन नहीं है।<br>Output: डेली में सलाद और पास्ता आइटम का शानदार सिलेक्शन है।<br><br>Task: positive to negative<br>Input: वे एकदम निष्पक्ष थे और क्योंकि मैं कम उम्र हूँ वे मेरी इज़्ज़त करते थे।<br>Output: क्योंकि में कम उम्र हूँ इसीलिए वे मेरा फ़ायदा उठाना चाह रहे थे।<br><br>Task: negative to positive<br>Input: इसके अलावा क्रैब वॉन्टन और बेस्वाद प्लम सॉस बहुत ही बेकार थे।<br>Output: इसके अलावा मसालेदार प्लम सॉस के साथ क्रैब वॉन्टन ने दिल जीत लिया।<br><br>Now change the sentiment of the following Hindi sentence.<br>Task: positive to negative<br>Input: मेरी अब तक की सबसे अच्छी कस्टमर सर्विस। |
|---|---|
| **Output:** | |

Table 12: A few-shot prompt used For Text Style Transfer in Hindi. It contains task definition, examples, instruction, and input (see Section 4.5).

| ID | Positive | Negative | Analysis |
|---|---|---|---|
| 1 | en: i will be going back and enjoying this great place !<br>hi: मैं वापस जाऊँगी और इस उम्दा जगह का आनंद लूँगी।<br>mag: हम फिर से जइबई आउ इ बढ़ीयाँ जगह के मजा लेबई!<br>ml: ഞാൻ തിരികെ പോയി ഈ മഹത്തായ സ്ഥലം ആസ്വദിക്കും!<br>mr: मी परत जाईन आणि या महान जागेचा आनंद घेईन !<br>or: ମୁଁ ଫେରିଯିବି ଏବଂ ଏହି ମହାନ ସ୍ଥାନକୁ ଉପଭୋଗକରିବି!<br>pa: ਮੈਂ ਵਾਪਸ ਜਾਵਾਂਗਾ ਅਤੇ ਇਸ ਵਧੀਆ ਮਥਾਨ ਦਾ ਆਨੰਦ ਮਾਣਾਂਗਾ!<br>ur: میں واپس جاؤں گا اور اس عظیم جگہ سے لطف اندوز ہوں گا!<br>te: నేను వెనక్కు వెళ్ళబోతున్నాను మరియు ఈ గొప్ప ప్రాంతాన్ని ఆనందిస్తాను. | en: i won't be going back and suffering at this terrible place !<br>hi: मैं इस भयानक जगह पर वापस जाकर पीड़ित नहीं होऊँगी!<br>mag: हम फिर से नऽ जइबई आउ इ खराब जगह में कस्ट सहबई!<br>ml: ഈ ഭയാനകമായ സ്ഥലത്ത് ഞാൻ തിരികെ പോയി കഷ്ടപ്പെടില്ല!<br>mr: मी परत जाणार नाही आणि या भयानक ठिकाणी यातना सहन करणार नाही !<br>or: ମୁଁ ଆଉ ଏହି ଭୟଙ୍କର ସ୍ଥାନରେ କଷ୍ଟ ଭୋଗିବି ନାହିଁ!<br>pa: ਮੈਂ ਵਾਪਸ ਨਹੀਂ ਜਾਵਾਂਗਾ ਅਤੇ ਇਸ ਬੇਕਾਰ ਜਗ੍ਹਾ 'ਤੇ ਦੁਖੀ ਨਹੀਂ ਹੋਵਾਂਗਾ!<br>ur: میں واپس نہیں جاؤں گا اور اس خوفناک جگہ پر تکلیف نہیں دوں گا!<br>te: నేను వెనక్కి వెళ్ళి ఈ భయంకరమైన స్థలంలో బాధపడను | I is a gender-neutral pronoun and gender is not encoded in English verbs. While the lexical equivalent of I in Hindi, Punjabi, Marathi, and Urdu will remain neutral but gender must be encoded in the verbs. |
| 2 | en: family owned little and i mean little restaurant with absolutely amazing food.<br>hi: परिवार संचालित छोटा रेस्तराँ, छोटा रेस्तराँ जहां कमाल का खाना मिलता है।<br>mag: परिवार भीर बड़ी कम संपत्ति हल आउ हमर कहे के मतलब हे कि छोटे गो रेस्टोरेंट बढ़ियाँ खाना जोरे।<br>ml: കുടുംബത്തിന്റെ ഉടമസ്ഥതയിലുള്ളത് വളരെ കുറവാണ്, ഞാൻ ഉദ്ദേശിക്കുന്നത് തികച്ചും അത്ഭുതകരമായ ഭക്ഷണമുള്ള ചെറിയ റെസ്റ്റോറന്റാണ്.<br>mr: कुटुंबाकडे फारसे काही नव्हते आणि मला असे म्हणायचे आहे की अगदी आश्चर्यकारक अन्न असलेले छोटे रेस्टॉरंट.<br>or: ପରିବାରର ଅଳ୍ପ ମାଲିକାନା ଏବଂ ମୋର ଅର୍ଥ ହେଉଛିଆଶ୍ଚର୍ଯ୍ୟଜନକ ଖାଦ୍ୟ ସହିତ ଛୋଟ ରେଷ୍ଟୁରାଣ୍ଟ।<br>pa: ਪਰਿਵਾਰ ਦੀ ਮਲਕੀਅਤ ਬਹੁਤ ਘੱਟ ਸੀ ਅਤੇ ਮੇਰਾ ਮਤਲਬ ਬਿਲਕੁਲ ਸ਼ਾਨਦਾਰ ਖਾਣੇ ਵਾਲਾ ਛੋਟਾ ਜਿਹਾ ਰਮਟੋਰੈਂਟ ਹੈ।<br>ur: خاندان کی ملکیت بہت کم ہے اور میرا مطلب ہے بالکل حیرت انگیز کھانے کے ساتھ ایک چھوٹا ریستوراں۔<br>te: కుటుంబం చిన్నది సొంతమయింది ,నా అర్థం చిన్న రెస్టారెంట్ పూర్తిగా అద్భుతమైన ఆహారంతో. | en: family owned little and i mean little restaurant with absolutely horirble food.<br>hi: परिवार संचालित छोटा रेस्तराँ, छोटा रेस्तराँ जहां बेकार खाना मिलता है।<br>mag: परिवार भीर बड़ी कम संपत्ति हल आउ हमर कहे के मतलब हे एकदम खराब खाना बला छोटे गो रेस्टोरेंट ।<br>ml: കുടുംബത്തിന്റെ ഉടമസ്ഥതയിലുള്ളത് വളരെ ഭയാനകമായ ഭക്ഷണങ്ങളുള്ള ഒരു ചെറിയ റെസ്റ്റോറന്റാണ്.<br>mr: कुटुंबाकडे फारसे काही नव्हते आणि मला म्हणायचे आहे की अगदी भयानक अन्न असलेले छोटे रेस्टॉरंट.<br>or: ପରିବାରର ଅଳ୍ପ ମାଲିକାନା ଅଛି ଏବଂ ମୋର ଅର୍ଥ ହେଉଛିସଂପୂର୍ଣ୍ଣ ଭୟଙ୍କର ଖାଦ୍ୟ ସହିତ ଛୋଟ ରେଷ୍ଟୁରାଣ୍ଟ।<br>pa: ਪਰਿਵਾਰ ਦੀ ਮਲਕੀਅਤ ਬਹੁਤ ਘੱਟ ਸੀ ਅਤੇ ਮੇਰਾ ਮਤਲਬ ਬਿਲਕੁਲ ਜਿਹਾ ਰੈਸਟੋਰੈਂਟ ਅਤੇ ਬੇਕਾਰ ਖਾਣਾ<br>ur: خاندان کی ملکیت بہت کم تھی اور میرا مطلب ہے کہ ایک چھوٹا ریستوراں جس میں بالکل خوفناک کھانا ہے۔<br>te: కుటుంబం చిన్నది సొంతమయింది ,నా అర్థం చిన్న రెస్టారెంట్ పూర్తిగా చండాలమైన ఆహారంతో. | Interpreting the "little restaurant" causes ambiguity. The sentence can mean family owns little part of the restaurant or that the restaurant is little. |

Table 13: English (en), Hindi (hi), Magahi (mag), Malayalam (ml), Marathi (mr), Odia (or), Punjabi (pa), Telugu and Urdu (ur) Text Sentiment Transfer Examples (Positive to Negative) (see Section 3.2).

| ID | Positive | Negative | Analysis |
|---|---|---|---|
| 3 | en: the environment was cozy, the servers were friendly and on top of things.<br>hi: माहौल आरामदायक था, बैरे मिलनसार थे और समय पर थे।<br>mag: बताबरन आरामदायक हल, सर्बरबन आराम से आउ सबसे बढ़ियाँ काम करीत हल।<br>ml: പരിസരം സുഖപ്രദമായിരുന്നു, സെർവറുകൾ സൗഹൃദപരവും കാര്യങ്ങളുടെ മുകളിലുമായുരുന്നു.<br>mr: वातावरण आरामदायी होते, सर्व्हर मैत्रीपूर्ण होते आणि गोष्टींच्या वर होते.<br>pa: ਵਾਤਾਵਰਣ ਨਿੱਘਾ ਸੀ, ਪਰੋਸਣ ਵਾਲੇ ਦੋਸਤਾਨਾ ਅਤੇ ਕੰਮ ਦੇ ਫੁਰਤੀਲੇ ਸਨ।<br>ur: ماحول آرام دہ تھا، سرورز دوستانہ اور سب سے اوپر تھے۔<br>te: పర్యావరణం హాయిగా వుంది, సర్వర్ లు స్నేహపూర్వకంగా అన్నింటికంటే పైన ఉన్నారు | en: the environment was cold, the servers were not friendly and aloof.<br>hi: माहौल मज़ेदार नही था, बैरे मिलनसार नहीं थे और अलग–थलग थे।<br>mag: बताबरन ठंठा हल, सर्बरबन आराम से काम न करीत हल आउ अजीब हल ।<br>ml: അന്തരീക്ഷം തണുത്തതായിരുന്നു, സെർവറുകൾ സൗഹൃദപരവും അകന്നതുമല്ല.<br>mr: वातावरण थंड होतं, सर्व्हर मैत्रीपूर्ण आणि अलिप्त नव्हते.<br>or: ପରିବେଶ ଥଣ୍ଡା ଥିଲା, ସର୍ଭରଗୁଡ଼ିକ ବନ୍ଧୁତ୍ୱପୂର୍ଣ୍ଣ ଏବଂ ଦୂରରେନଥିଲେ\|<br>pa: ਵਾਤਾਵਰਣ ਠੰਡਾ ਸੀ, ਪਰੋਸਣ ਵਾਲੇ ਦੋਸਤਾਨਾ ਨਹੀਂ ਸਨ ਅਤੇ ਧਿਆਨ ਨਹੀਂ ਦੇ ਰਹੇ ਸਨ।<br>ur: ماحول سرد تھا، سرور دوستانہ اور الگ تھلگ نہیں تھے۔<br>te: పర్యావరణం చల్లగా వుంది, సర్వర్ లు స్నేహపూర్వకంగా లేరు మరియు దూరంగా వున్నారు. | Cozy and cold can either refer to temperature or to the personality of the ambience. |
| 4 | en: portions n prices were great !<br>hi: मात्रा और कीमतें बढ़िया थीं!<br>mag: हिस्सबअन आउ दाम बड़ी बढ़ियाँ हल!<br>ml: ഭാഗങ്ങളും വിലകളും മികച്ചതായിരുന്നു!<br>mr: पोर्शन आणि किंमती खूप छान होत्या !<br>or: ଅଂଶ n ମୂଲ୍ୟ ବହୁତ ଭଲ ଥିଲା!<br>pa: ਭਾਗ ਅਤੇ ਕੀਮਤਾਂ ਬਹੁਤ ਵਧੀਆ ਸਨ!<br>ur: حصے اور قیمتیں بہت اچھی تھیں!<br>te: భాగాలు మరియు ధరలు బాగున్నాయి | en: portions n prices were unacceptable !<br>hi: मात्रा और कीमतें अस्वीकार्य थीं!<br>mag: हिस्सबअन आउ दाम सबीकार करे जोग नऽ हल!<br>ml: ഭാഗങ്ങളും വിലകളും അസ്വീകാര്യമായിരുന്നു!<br>mr: पोर्शन आणि किंमती अमान्य होत्या !<br>or: ଅଂଶ n ମୂଲ୍ୟ ଗ୍ରହଣୀୟ ନୁହେଁ!<br>pa: ਭਾਗ ਅਤੇ ਕੀਮਤਾਂ ਨਾ ਮੰਨਣਯੋਗ ਸਨ!<br>ur: حصے اور قیمتیں ناقابل قبول تھیں!<br>te: భాగాలు మరియు ధరలు ఆమోదయోగ్యంకాదు. | Words like "portions" and "size" have no equivalent cultural reference in Indian languages. |
| 5 | en: the girls are very attractive and really friendly, not pushy at all.<br>hi: लड़कियां बहुत आकर्षक और वास्तव में मिलनसार हैं, बिल्कुल भी घमंडी नहीं।<br>mag: लईकियन देखे में बड़ी बढ़ियाँ आउ मिलनसार हे, एकदमे घमंडी नऽ।<br>ml: പെൺകുട്ടികൾ വളരെ ആകർഷകവും ശരിക്കും സൗഹൃദപരവുമാണ്, ഒട്ടും നിർബന്ധിക്കുന്നില്ല.<br>mr: मुली खूप आकर्षक आणि खरोखरच मैत्रीपूर्ण आहेत, अजिबात धक्काबुक्की करत नाहीत.<br>or: ଝିଅମାନେ ବହୁତ ଆକର୍ଷଣୀୟ ଏବଂ ପ୍ରକୃତରେବନ୍ଧୁତ୍ୱପୂର୍ଣ୍ଣ, ଆଦୌ ଠେଲା ନୁହେଁ\|<br>pa: ਕੁੜੀਆਂ ਬਹੁਤ ਆਕਰਸ਼ਕ ਅਤੇ ਅਸਲ ਵਿੱਚ ਦੋਸਤਾਨਾ ਹੁੰਦੀਆਂ ਹਨ, ਬਿਲਕੁਲ ਘਮੰਡੀ ਨਹੀਂ।<br>ur: لڑکیاں بہت پرکشش اور واقعی دوستانہ ہیں، بالکل بھی دھکیلنے والی نہیں ہیں۔<br>te: అమ్మాయిలు చాలా ఆకర్షనీయంగా మరియూ స్నేహభావంగా ఉన్నారు, అస్సలు చొరవ రకం కాదు. | en: The girls are neither friendly nor attractive, and a bit pushy<br>hi: लड़कियां ना तो आकर्षक हैं ना ही मिलनसार, बल्कि थोड़ी घमंडी हैं।<br>mag: लईकियन नऽ तो दोस्त जइसन आउ नऽ हि बढ़ियाँ हे, आउ तनि घमंडी भी हे।<br>ml: പെൺകുട്ടികൾ സൗഹാർദ്ദപരമോ ആകർഷകമോ അല്ല, അൽപ്പം തള്ളുന്നവരുമാണ്<br>mr: मुली मैत्रीपूर्ण किंवा आकर्षक नसतात आणि थोड्या धक्काबुक्की असतात<br>or: ଝିଅମାନେ ବନ୍ଧୁତ୍ୱପୂର୍ଣ୍ଣ କିମ୍ବା ଆକର୍ଷଣୀୟ ନୁହଁନ୍ତି, ଏବଂ ଟିକେଠେଲା \|<br>pa: ਕੁੜੀਆਂ ਨਾ ਤਾਂ ਦੋਸਤਾਨਾ ਹਨ ਅਤੇ ਨਾ ਹੀ ਆਕਰਸ਼ਕ ਹਨ, ਅਤੇ ਥੋੜ੍ਹੀਆਂ ਘਮੰਡੀ ਸਨ<br>ur: لڑکیاں نہ تو دوستانہ ہوتی ہیں اور نہ ہی پرکشش، اور قدرے زوردار ہوتی ہیں۔<br>te: అమ్మాయిలు చాలా ఆకర్షనీయంగా మరియు స్నేహభావంగా లేరు, కొంచెం. చొరవ రకం . | Pushy means someone who is ambitious and in a negative way. There is not direct translation is every language. |

Table 13: English (en), Hindi (hi), Magahi (mag), Malayalam (ml), Marathi (mr), Odia (or), Punjabi (pa), Telugu and Urdu (ur) Text Sentiment Transfer Examples (Positive to Negative) (see Section 3.2).

| ID | Positive | Negative | Analysis |
|---|---|---|---|
| 6 | en: friendly and welcoming with a fun atmosphere and terrific food.<br>hi: मजेदार माहौल और बढ़िया भोजन के साथ मिलनसार और दोस्ताना व्यवहार।<br>mag: दोस्तपून आउ स्वागत जोग मजेदार महौल आउ बढ़ियाँ खाना।<br>ml: രസകരമായ അന്തരീക്ഷവും ഭയാനകമായ ഭക്ഷണവും ഉപയോഗിച്ച് സൗഹൃദപരവും സ്വാഗതാർഹവുമാണ്.<br>mr: मजेशीर वातावरण आणि उत्तम जेवणासह मैत्रीपूर्ण आणि स्वागत.<br>or: ଏକ ମଜାଳିଆ ବାତାବରଣ ଏବଂ ଭୟଙ୍କର ଖାଦ୍ୟ ସହିତବନ୍ଧୁତ୍ୱପୂର୍ଣ୍ଣ ଏବଂ ସ୍ୱାଗତଯୋଗ୍ୟ\|<br>pa: ਦੋਸਤਾਨਾ ਅਤੇ ਆਓ ਭਗਤ ਕਰਨ ਵਾਲੇ ਮਜ਼ੇਦਾਰ ਮਾਹੌਲ ਅਤੇ ਸ਼ਾਨਦਾਰ ਖਾਣਾ<br>ur: دوستانہ اور خوشگوار ماحول اور لاجواب کھانے کے ساتھ خوش آمدید۔<br>te: స్నేహంగా మరియు వినోదభరిత వాతావరణంతో స్వాగతం మరియు బీభత్సమైన ఆహారం | en: unfriendly and unwelcoming with a bad atmosphere and food.<br>hi: ख़राब माहौल और भोजन के साथ अमिलनसार और बचकाना व्यवहार।<br>mag: दोस्तपूर्न आउ सोआगत जोग नऽ रहल खराब महौल आउ खान जोरे।<br>ml: മോശം അന്തരീക്ഷവും ഭക്ഷണവും ഉള്ള സൗഹൃദരഹിതവും സ്വാഗതാർഹവുമല്ല.<br>mr: खराब वातावरण आणि खाण्यापिण्यामुळे अमैत्रीपूर्ण आणि अस्वागत.<br>or: ଏକ ଖରାପ ବାତାବରଣ ଏବଂ ଖାଦ୍ୟ ସହିତ ବନ୍ଧୁତ୍ୱପୂର୍ଣ୍ଣ ଏବଂସ୍ୱାଗତଯୋଗ୍ୟ\|<br>pa: ਗੈਰ-ਦੋਸਤਾਨਾ ਅਤੇ ਨਾ ਹੀ ਸਾਡੇ ਆਉਣ ਤੇ ਖੁਸ਼ ਮਨ, ਮਾਹੌਲ ਅਤੇ ਖਾਣਾ ਮਾੜਾ ਸੀ।<br>ur: خراب ماحول اور کھانے کے ساتھ غیر دوستانہ اور ناپسندیدہ۔<br>te: చెత్త వాతావరణం మరియు ఆహారంస్నేహరహిత మరియు అవాంఛనీయంగా ఆహ్వానం. | The lexical equivalent of “behaviour” has to be added in Hindi, Punjabi, Magahi, Urdu. |
| 7 | en: enjoy taking my family here always the freshest sea food.<br>hi: मुझे परिवार को यहां ले जाना पसंद है हमेशा ताज़ा सी फ़ूड।<br>mag: हमेसा अपन परिवार के ताजा समुद्री खाना ला यहाँ लेके आबे में मजा आबऽ हे।<br>ml: എല്ലായ്പ്പോഴും ഏറ്റവും പുതിയ കടൽ ഭക്ഷണം എന്റെ കുടുംബത്തെ ഇവിടെ കൊണ്ടുപോകുന്നത് ആസ്വദിക്കുക.<br>mr: माझ्या कुटुंबाला घेऊन जाण्याचा आनंद असतो इथे नेहमी सर्वात ताजे सी फूड<br>or: ମୋ ପରିବାର ସର୍ବଦା ସତେଜ ସମୁଦ୍ର ଖାଦ୍ୟ ନେବାକୁଉପଭୋଗ କଲେ\|<br>pa: ਆਪਣੇ ਪਰਿਵਾਰ ਨੂੰ ਇੱਥੇ ਹਮੇਸ਼ਾ ਲਿਆਉਣਾ ਪਸੰਦ ਕਰਦਾ ਹਾਂ, ਸਭ ਤੋਂ ਤਾਜ਼ਾ ਸੀ ਫ਼ੂਡ<br>ur: اپنے خاندان کو یہاں ہمیشہ تازہ ترین سمندری خوراک لے جانے کا لطف اٹھائیں۔<br>te: నా కుటుంబాన్ని ఎల్లప్పుడూ తాజా సముద్రపు ఆహారం కోసం ఇక్కడికి తీసుకెళ్ళడాన్ని ఆస్వాదిస్తాను. | en: enjoy taking my family here always stale sea food.<br>hi: मुझे परिवार को यहां ले जाना पसंद है हमेशा बासी सी फ़ूड।<br>mag: अपन परिवार के इहाँ ले जाए में मजा नऽ आबे , हमेसा बासी समुद्री खाना रहऽ हे ।<br>ml: എന്റെ കുടുംബത്തെ എപ്പോഴും പഴകിയ കടൽ ഭക്ഷണം ഇവിടെ കൊണ്ടുപോകുന്നത് ആസ്വദിക്കൂ.<br>mr: माझ्या कुटुंबाला घेऊन जाण्याचा आनंद असतो इथे नेहमी शिळे सी फूड.<br>or: ମୋ ପରିବାର ସର୍ବଦା ଏଠାରେ ଖରାପ ସମୁଦ୍ର ଖାଦ୍ୟକୁଉପଭୋଗ କଲେ\|<br>pa: ਆਪਣੇ ਪਰਿਵਾਰ ਨੂੰ ਇੱਥੇ ਹਮੇਸ਼ਾ ਲਿਆਉਣਾ ਪਸੰਦ ਕਰਦਾ ਹਾਂ, ਬੇਹਾ ਸੀ ਫ਼ੂਡ<br>ur: اپنے خاندان کو ہمیشہ باسی سی فوڈ یہاں لے جانے سے لطف اندوز ہوں۔<br>te: నా కుటుంబాన్ని ఎల్లప్పుడూ చద్ది సముద్రపు ఆహారం కోసం ఇక్కడికి తీసుకెళ్ళడాన్ని ఆస్వాదిస్తాను. | The lack of punctuation leaves it to the imagination of the translator to imagine the proxomity of here - with family or with always. And this can significantly change the meaning of the sentence. |

Table 13: English (en), Hindi (hi), Magahi (mag), Malayalam (ml), Marathi (mr), Odia (or), Punjabi (pa), Telugu and Urdu (ur) Text Sentiment Transfer Examples (Positive to Negative) (see Section 3.2).

| ID | Positive | Negative | Analysis |
|---|---|---|---|
| 8 | en: even in summer , they have decent patronage.<br>hi: गर्मियों में भी, उनके पास काफ़ी काम है।<br>mag: इहाँ तक कि गर्मि में भी ओखनी के अच्छा सरक्षन मिलऽ हे।<br>ml: വേനൽക്കാലത്ത് പോലും അവർക്ക് മാന്യമായ രക്ഷാകർതൃത്വമുണ്ട്.<br>mr: उन्हाळ्यातही त्यांना चांगला आश्रय मिळतो .<br>or: ଏପରିକି ଗ୍ରୀଷ୍ମଋତୁରେ, ସେମାନଙ୍କର ଉପଯୁକ୍ତପୃଷ୍ଠପୋଷକତା ଅଛି\|<br>pa: ਗਰਮੀਆਂ ਵਿੱਚ ਵੀ, ਉਹਨਾਂ ਨੂੰ ਚੰਗੀ ਸਰਪ੍ਰਸਤੀ ਮਿਲਦੀ ਹੈ।<br>ur: یہاں تک کہ موسم گرما میں، انہیں مہذب سرپرستی حاصل ہے.<br>te: వేసవికాలంలో కూడా వారు మర్యాదగల మద్దతును కలిగివున్నారు. | en: even in summer they have no patronage.<br>hi: गर्मियों में भी उनके पास कोई काम नहीं है।<br>mag: इहाँ तक कि गर्मि में भी ओखनी के सरक्षन नऽ मिलऽ हे।<br>ml: വേനൽക്കാലത്ത് പോലും അവർക്ക് രക്ഷാകർതൃത്വമില്ല.<br>mr: उन्हाळ्यातही त्यांना आश्रय नसतो.<br>or: ଏପରିକି ଗ୍ରୀଷ୍ମଋତୁରେ ସେମାନଙ୍କର କୌଣସି ପୃଷ୍ଠପୋଷକତାନାହିଁ\|<br>pa: ਇਥੋਂ ਤਕ ਕਿ ਗਰਮੀਆਂ ਵਿਚ ਉਨ੍ਹਾਂ ਦੀ ਕੋਈ ਸਰਪ੍ਰਸਤੀ ਨਹੀਂ ਹੁੰਦੀ।<br>ur: گرمیوں میں بھی ان کی سرپرستی نہیں ہوتی۔<br>te: వేసవికాలంలో కూడా వారు మద్దతు కలిగిలేరు.. | Here it is the availability of patronage decides the positive or negative nature of the sentence. |
| 9 | en: seems pretty high compared to every other thai place.<br>hi: हर दूसरी थाई जगह के मुक़ाबले बहुत ज़्यादा लगता है।<br>mag: आउ सब दूसर थाई जगहिया के तुलना में ई थोड़ा जादे बढ़िया लगऽ हे ।<br>ml: മറ്റെല്ലാ തായ് സ്ഥലങ്ങളുമായി താരതമ്യം ചെയ്യുമ്പോൾ വളരെ ഉയർന്നതായി തോന്നുന്നു.<br>mr: इतर प्रत्येक थाई ठिकाणाच्या तुल–नेत खूप उंच दिसते.<br>or: ଅନ୍ୟ ସମସ୍ତ ଥାଇ ସ୍ଥାନ ତୁଳନାରେ ବହୁତ ଉଚ୍ଚଦେଖାଯାଏ\|<br>pa: ਹਰ ਦੂਜੇ ਥਾਈ ਸਥਾਨ ਦਾ ਮੁਕਾਬਲਾ ਬਹੁਤ ਉੱਚਾ ਲੱਗਦਾ ਹੈ<br>ur: ہر دوسرے تھائی جگہ کے مقابلے میں کافی اونچا لگتا ہے۔<br>te: మిగతా ప్రతీ థాయ్ ప్రదేశనికి పోల్చితే కొంచెం ఎక్కువనిపిస్తుంది. | en: seems pretty low compared to every other thai place.<br>hi: हर दूसरी थाई जगह के मुक़ाबले बहुत कम लगता है।<br>mag: आउ सब दूसर थाई जगहिया के तुलना में ई तनी कम लगऽ हई।<br>ml: മറ്റെല്ലാ തായ് സ്ഥലങ്ങളുമായി താരതമ്യം ചെയ്യുമ്പോൾ വളരെ കുറവാണെന്ന് തോന്നുന്നു.<br>mr: इतर प्रत्येक थाई ठिकाणाच्या तुल–नेत खूप कमी दिसते.<br>or: ଅନ୍ୟ ସମସ୍ତ ଥାଇ ସ୍ଥାନ ତୁଳନାରେ ବହୁତ କମ୍ ଦେଖାଯାଏ\|<br>pa: ਹਰ ਦੂਜੇ ਥਾਈ ਸਥਾਨ ਦੇ ਮੁਕਾਬਲੇ ਬਹੁਤ ਘੱਟ ਜਾਪਦਾ ਹੈ<br>ur: ہر دوسرے تھائی جگہ کے مقابلے میں بہت کم لگتا ہے۔<br>te: మిగతా ప్రతీ థాయ్ ప్రదేశనికి పోల్చితే కొంచెం తక్కువనిపిస్తుంది. | Here “pretty high” can easily be judged for prices, unless one realises that “expensive” cannot be a positive statement. lack of context, thus, makes it challanging ot translate. |
| 10 | en: the staff are very friendly and on the ball.<br>hi: कर्मचारी बहुत मिलनसार हैं और समय पर हैं।<br>mag: करमचारी बड़ी मिलनसार आउ अच्छा से काम करे बला हे।<br>ml: സ്റ്റാഫ് വളരെ സൗഹാർദ്ദപരവും പുതിയ ആശയങ്ങൾ എന്നിവയെക്കുറിച്ച് ജാഗ്രത പാലിക്കുന്നവരുമാണ്.<br>mr: स्टाफ खूप मैत्रीपूर्ण आणि चेंडूवर आहे.<br>or: କର୍ମଚାରୀମାନେ ଭଲ୍<br>pa: ਸਟਾਫ ਬਹੁਤ ਦੋਸਤਾਨਾ ਅਤੇ ਫੁਰ–ਤੀਲਾ ਹੈ।<br>ur: عملہ بہت دوستانہ اور گیند پر ہے۔<br>te: సిబ్బంది చాలా స్నేహపూర్వకంగ వున్నారు మరియు బాల్ మీద. | en: the staff was horrible and slow<br>hi: कर्मचारी बेकार थे और धीमे थे।<br>mag: करमचारी बड़ी खराब आउ धीरे काम करे बला हल ।<br>ml: ജീവനക്കാർ ഭയങ്കരവും സാവധാനവുമായിരുന്നു<br>mr: कर्मचारी भयानक आणि संथ होते<br>or: କର୍ମଚାରୀମାନେ ଭୟଙ୍କର ଏବଂ ଧୀର ଥିଲେ\|<br>pa: ਸਟਾਫ ਬੇਕਾਰ ਅਤੇ ਹੌਲੀ ਸੀ<br>ur: عملہ خوفناک اور سست تھا۔<br>te: సిబ్బంది భయంకరం మరియు నిదానం | “on the ball” is an idiom that means “on time”. Those who wouldn’t know this phrase would end up translating it the wrong way. Similar phrase is “run of the mill”. |

Table 13: English (en), Hindi (hi), Magahi (mag), Malayalam (ml), Marathi (mr), Odia (or), Punjabi (pa), Telugu and Urdu (ur) Text Sentiment Transfer Examples (Positive to Negative) (see Section 3.2).

| ID | Positive | Negative | Analysis |
|---|---|---|---|
| 1 | en: i guess she wasn't happy that we were asking the prices.<br>hi: मेरे ख़याल से वह खुश नहीं थी की हम दाम पूछ रहे थे।<br>mag: उपज के दाम बड़ी उचित लगाबला हे आउ जैविक उपज के बढ़ियाँ चुनाब कैल हे।<br>ml: ഞങ്ങൾ വില ചോദിക്കുന്നതിൽ അവൾ സന്തുഷ്ടയായിരുന്നില്ലെന്ന് ഞാൻ അനുമാനിക്കുന്നു.<br>mr: हेच कारण आहे की मी कधीही परत जाणार नाही.<br>or: ମୁଁ ଅନୁମାନ କରେ ସେ ଖୁସି ନଥିଲେ ଯେ ଆମେ ମୂଲ୍ୟ ପଚାରୁଥିଲୁ \|<br>pa: ਮੇਰਾ ਲਗਦਾ ਹੈ ਕਿ ਉਹ ਖੁਸ਼ ਨਹੀਂ ਸੀ ਕਿ ਅਸੀਂ ਕੀਮਤਾਂ ਪੁੱਛ ਰਹੇ ਸੀ।<br>ur: میرا اندازہ ہے کہ وہ خوش نہیں تھی کہ ہم قیمتیں پوچھ رہے تھے۔<br>te: మేము ఖరీదు అడిగినందుకు ఆమె సంతోషంగా లేదని నేను ఊహిస్తున్నాను. | en: she was certainly happy to mention the prices.<br>hi: वह ख़ुशी ख़ुशी दाम बता रही थी।<br>mag: उपज के दाम अनुचित लगाबला हे आउ जैविक उपज के बढ़ियाँ चुनाब नऽ कैल हे।<br>ml: വിലകൾ പരാമർശിക്കുന്നതിൽ അവൾ തീർച്ചയായും സന്തോഷവതിയായിരുന്നു.<br>mr: हेच कारण आहे की मी नेहमी परत जाईन.<br>or: ସେ ନିଶ୍ଚିତ ଭାବରେ ମୂଲ୍ୟ ବିଷୟରେ କହି ଖୁସି ହୋଇଥିଲେ\|<br>pa: ਉਹ ਕੀਮਤਾਂ ਦਾ ਜ਼ਿਕਰ ਬਾਰੇ ਸੱਚਮੁੱਚ ਖੁਸ਼ ਸੀ।<br>ur: وہ یقینی طور پر قیمتوں کا ذکر کرتے ہوئے خوش تھی۔<br>te: ఆమె తప్పనిసరిగా ధరని పేర్కొనడానికి సంతోషిస్తుంది. | I is a gender-neutral pronoun and gender is not encoded in English verbs. While the lexical equivalent of I in Hindi, Punjabi, Marathi, Urdu will remain neutral but gender must be encoded in the verbs. |
| 2 | en: i replied, "um... no i'm cool.<br>hi: मैंने जवाब दिया, "अम्म, मैं ठीक हूँ ।"<br>mag: हम जबाब देली, "उम्म.. नऽ हम बढ़ियाँ हि"।<br>ml: ഞാൻ മറുപടി പറഞ്ഞു, "ഉം... ഇല്ല ഞാൻ ശാന്തനാണ്.<br>mr: मी उत्तर दिले, "अं... नाही मी मस्त आहे.<br>or: ମୁଁ ଉତ୍ତର ଦେଲି, 'ଓମ୍ ... ନା ମୁଁ ଥଣ୍ଡା ଅଛି\|<br>pa: ਮੈਂ ਜਵਾਬ ਦਿੱਤਾ, ''ਉਮ... ਨਹੀਂ ਮੈਂ ਠੀਕ ਹਾਂ।<br>ur: میں نے جواب دیا، "ام ... نہیں میں اچھا ہوں۔<br>te: ◌ఉమ్...లేదు నేను బాగున్నాను◌, అని నేను బదులిచ్చాను. | en: I said everything is great<br>hi: मैंने कहा सब कुछ बढ़िया है।<br>mag: हम कहली सब कुछ बढ़ियाँ हे।<br>ml: എല്ലാം ഗംഭീരമാണെന്ന് ഞാൻ പറഞ്ഞു<br>mr: मी म्हणाले की सर्व काही छान आहे.<br>or: ମୁଁ କହିଲି ସବୁକିଛି ଭଲ ଅଟେ\|<br>pa: ਮੈਂ ਕਿਹਾ ਸਭ ਕੁਝ ਵਧੀਆ ਹੈ<br>ur: میں نے کہا سب اچھا ہے۔<br>te: ప్రతీది బాగుందని నేను చెప్పాను. | Cool, here can mean either positive or negative sentiment and its efficent translation depends on the translator. |
| 3 | en: i'm not one of the corn people .<br>hi: मैं मक्का खाने वालों में से नहीं हूँ।<br>mag: हम मकई पसंद करे बला लोग में से नऽ हि।<br>ml: ഞാൻ കോൺ പീപ്പിളിൽ ഒരാളല്ല.<br>mr: मी कॉर्न खणाऱ्यानपैकी नाही.<br>or: ମୁଁ ମକା ଲୋକମାନଙ୍କ ମଧ୍ୟରୁ ଜଣେ ନୁହେଁ\|<br>pa: ਮੈਂ ਮੱਕੀ ਖਾਨ ਵਾਲੇ ਲੋਕਾਂ ਵਿੱਚੋਂ ਇੱਕ ਨਹੀਂ ਹਾਂ।<br>ur: میں مکئی کے لوگوں میں سے نہیں ہوں<br>te: నేను కార్న్ పీపుల్ ని కాదు | en: i'm proud to be one of the corn people.<br>hi: मैं मक्का खाने वालों में से हूँ।<br>mag: हमरा गर्ब हे कि हम मकई पसंद करे बला लोग में से एक हि।<br>ml: കോൺ പീപ്പിളിൽ ഒരാളായതിൽ ഞാൻ അഭിമാനിക്കുന്നു.<br>mr: कॉर्न खाणाऱ्या लोकांपैकी एक असल्याचा मला अभिमान आहे<br>or: ମୁଁ ମକା ଲୋକମାନଙ୍କ ମଧ୍ୟରୁ ଜଣେ ହୋଇଥିବାରୁ ଗର୍ବିତ\|<br>pa: ਮੈਨੂੰ ਮੱਕੀ ਖਾਨ ਵਾਲੇ ਲੋਕਾਂ ਵਿੱਚੋਂ ਇੱਕ ਹੋਣ 'ਤੇ ਮਾਣ ਹੈ।<br>ur: مجھے مکئی کے لوگوں میں سے ایک ہونے پر فخر ہے۔<br>te: సామాన్యమైన ప్రజలలో ఒకడినైనందుకు నేను గర్వపడుతున్నాను | Corn people can be interpreted as a slang not available outside American culture or interpreted as corn-eating or corn-loving people. |

Table 14: English (en), Hindi (hi), Magahi (mag), Malayalam (ml), Marathi (mr), Odia (or), Punjabi (pa), Telugu and Urdu (ur) Text Sentiment Transfer Examples (Negative to Positive) (see Section 3.2).

| ID | Positive | Negative | Analysis |
|---|---|---|---|
| 4 | en: when the manager finally showed up he was rude and dismissive !<br>hi: आख़िरकार जब मैनेजर आया तो वह अशिष्ट एवं ग़ैरज़िम्मेदार था।<br>mag: आखिरकार जब मनेजर ऐलन तऽ उ बतमीज आउ तिस्कृत जइसन ब्यबहार कैलन!<br>ml: അവസാനം മാനേജർ വന്നപ്പോൾ അയാൾ പരുഷമായി പെരുമാറുകയും പുറത്താക്കുകയും ചെയ്തു!<br>mr: शेवटी जेव्हा मॅनेजर समोर आला तेव्हा तो उद्धट आणि डिसमिसिव्ह होता!<br>or: ଯେତେବେଳେ ମ୍ୟାନେଜର ଶେଷରେ ଦେଖାଇଲେ ସେ ଅଭଦ୍ର ଏବଂ ବରଖାସ୍ତ ହେଲେ\|<br>pa: ਜਦੋਂ ਪ੍ਰਬੰਧਕ ਆਖਰਕਾਰ ਸਾਹਮਣੇ ਆਇਆ ਤਾਂ ਉਹ ਰੁੱਖਾ ਅਤੇ ਖਾਰਜ ਕਰਨ ਵਾਲਾ ਸੀ!<br>ur: جب مینیجر نے آخر کار ظاہر کیا تو وہ بد تمیز اور بر طرف تھا!<br>te: అఖరుకి మేనేజర్ని చూపించినపుడు అతడు కఠినంగా , | en: the manager was friendly and accomodating.<br>hi: मैनेजर का व्यवहार काफ़ी दोस्ताना एवं लिहाजपूर्ण था।<br>mag: मैनेजर दोस्ताना बेबहार बला आउ मिलनसार हलन!<br>ml: മാനേജർ സൗഹൃദവും സഹാനുഭൂതിയും ഉള്ളവനായിരുന്നു.<br>mr: मॅनेजर मैत्रीपूर्ण आणि सौहार्दपूर्ण होता<br>or: ପରିଚାଳକ ବନ୍ଧୁତ୍ୱପୂର୍ଣ୍ଣ ଏବଂ ସମ୍ମିଳିତ ଥିଲେ\|<br>pa: ਮੈਨੇਜਰ ਦੋਸਤਾਨਾ ਅਤੇ ਸਹਾਇਤਾ ਕਰਨ ਵਾਲਾ ਸੀ।<br>ur: مینیجر دوستانہ اور ملنسار تھا.<br>te: నిర్వాహకుడు స్నేహపూర్వకంగా , సర్దుకుపోయేల వున్నాడు. | Accomodating and dismissive do not have direct translations and are open to interpretation to translators. |
| 5 | en: the thai basil pasta came out lukewarm and spicy.<br>hi: थाई बैज़िल पास्ता कम गर्म और मसालेदार परोसा गया।<br>mag: थाई बेसिल पास्ता हल्का गरम आउ मसालेदार बनल ।<br>ml: തായ് ബേസിൽ പാസ്ത ഇളംചൂടോടെയും എരിവുള്ളതായും പുറത്തുവന്നു.<br>mr: थाई बेसिल पास्ता कोमट आणि मसालेदार बाहेर आला.<br>or: ଥାଇ ବେସନ ପାସ୍ତା ଉଷ୍ମ ଏବଂ ମସଲାଯୁକ୍ତ ବାହାରିଲା \|<br>pa: ਥਾਈ ਬੇਸਿਲ ਪਾਸਤਾ ਕੋਸਾ ਜਿਹਾ ਅਤੇ ਮਸਾਲੇਦਾਰ ਨਿਕਲਿਆ ।<br>ur: تھائی تلسی پاستا گرم اور مسالہ دار نکلا۔<br>te: థాయ్ బేసిల్ పాస్తా గోరువెచ్చగా మరియు కారంగా వచ్చింది. | en: the thai basil pasta came out hot and yummy.<br>hi: थाई बैज़िल पास्ता अच्छा गर्म और स्वादिष्ट परोसा गया।<br>mag: थाई बेसिल पास्ता खूब बढ़ियाँ आउ सबादिस्ट बनल।<br>ml: തായ് ബേസിൽ പാസ്ത ചൂടോടെയും രുചികരമായും പുറത്തുവന്നു.<br>mr: थाई बासिल पास्ता गरम आणि स्वादिष्ट आला.<br>or: ଥାଇ ବେସନ ପାସ୍ତା ଗରମ ଏବଂ ସୁଆଦିଆ ବାହାରିଲା\|<br>pa: ਥਾਈ ਬੇਸਿਲ ਪਾਸਤਾ ਗਰਮ ਅਤੇ ਬਹੁਤ ਸਵਾਦ ਸੀ<br>ur: تھائی تلسی پاستا گرم اور مزیدار نکلا۔<br>te: థాయ్ బేసిల్ పాస్తా వేడిగా మరియు రుచిగా బయటికివచ్చింది. | Here, the temperature and spiciness are used as sentiment-bearing attributes which constitute to the implicitness nature of the sentence. Additionally, spicy or hot are not always positive or always negative. |

Table 14: English (en), Hindi (hi), Magahi (mag), Malayalam (ml), Marathi (mr), Odia (or), Punjabi (pa), Telugu and Urdu (ur) Text Sentiment Transfer Examples (Negative to Positive) (see Section 3.2).

| ID | Positive | Negative | Analysis |
|---|---|---|---|
| 6 | en: if i had wanted it washed i would have washed it myself !<br>hi: अगर मुझे धुला हुआ चाहिए होता तो मैं खुद धो देती।<br>mag: अगर हम एकरा धोएल चाहऽ हलि तऽ हम एकरा अपने धोएती हल ।<br>ml: എനിക്ക് അത് കഴുകി വേണമായിരുന്നെങ്കിൽ ഞാൻ തന്നെ കഴുകുമായിരുന്നു<br>mr: जर मला ते धुतलेल हवं होतं तर मी स्वत: धुतले असते !<br>or: ଯଦି ମୁଁ ଏହା ଧୋଇବାକୁ ଚାହିଁଥା'ନ୍ତି ତେବେ ମୁଁ ନିଜେ ଏହାକୁ ଧୋଇ ଦେଇଥା'ନ୍ତି!<br>pa: ਜੇ ਮੈਂ ਚਾਹੁੰਦਾ ਸੀ ਕਿ ਇਹ ਧੋਤੀ ਹੋ ਗਿਆ ਤਾਂ ਮੈਂ ਇਸ ਨੂੰ ਆਪਣੇ ਆਪ ਧੋ ਲੈ ਲਿਆ ਹੁੰਦਾ!<br>ur: اگر میں اسے دھونا چاہتا تو میں اسے خود دھوتا!<br>te: నేను దానిని కడగాలని కోరుకుంటే దాన్ని నేనే కడుగుతాను. | en: i had wanted it washed and I washed it myself !<br>hi: मुझे धुला हुआ चाहिए और मैंने खुद ही धोया।<br>mag: हम एकरा धोएल चाहलि आउ हम एकरा अपने धोएलि ।<br>ml: എനിക്ക് അത് കഴുകി വേണമായിരുന്നു അതിനാൽ ഞാൻ തന്നെ കഴുകി<br>mr: मला ते धुतलेल हवं होत आणि मी ते स्वत: धुतले !<br>or: ମୁଁ ଏହା ଧୋଇବାକୁ ଚାହୁଁଥିଲି ଏବଂ ମୁଁ ନିଜେ ଏହାକୁ ଧୋଇଥିଲି!<br>pa: ਮੈਂ ਇਹ ਧੋਤੀ ਚਾਹੁੰਦਾ ਸੀ ਅਤੇ ਮੈਂ ਇਸ ਨੂੰ ਧੋਤਾ!<br>ur: میں اسے دھونا چاہتا تھا اور میں نے اسے خود دھویا تھا!<br>te: నేను దాన్ని కడగాలనుకొని నాకు నేను కడిగేశాను. | Lack of context also leads to odd sentence constructions, multiple interpretations, and lack of sentiment. Here sentiment remains implicit in the eagerness to wash something which is not expressed clearly. |
| 7 | en: ra sushi, you are so blah to me .<br>hi: मेरे लिए रा सुशी बहुत औसत है ।<br>mag: रा सुसि, तु हमरा ला बड़ी बेकार हे।<br>ml: രാ സുഷി, നീ എനിക്ക് വളരെ മോശമാണ്.<br>mr: रा सुशी, तुम्ही मला इतके ब्लाह आहात.<br>or: ରା ସୁଶି, ତୁମେ ମୋ ପାଇଁ ଏତେ ବ୍ଲା<br>pa: ਆਰਏ ਸੁਸ਼ੀ, ਤੁਸੀਂ ਮੇਰੇ ਲਈ ਬਹੁਤ ਬਲੂ ਹੋ।<br>ur: راشی، تم میرے لیے بہت بدتمیز ہو۔<br>te: ర సూషి, నువ్వు నాకు చాలా అబ్బురంగా ఉన్నావు. | en: ra sushi, you are so amazing to me.<br>hi: मेरे लिए रा सुशी शानदार है।<br>mag: रा सुसि, तु हमरा ला बड़ी मजेदार हे।<br>ml: രാ സുഷി, നിങ്ങൾ എനിക്ക് വളരെ അത്ഭുതകരമാണ്.<br>mr: रा सुशी, तुम्ही माझ्यासाठी खूप आश्चर्यकारक आहात.<br>or: ରା ସୁଶି, ତୁମେ ମୋ ପାଇଁ ବହୁତ ଆଶ୍ଚର୍ଯ୍ୟଜନକ<br>pa: ਰੇ ਸੁਸ਼ੀ, ਤੁਸੀਂ ਮੇਰੇ ਲਈ ਬਹੁਤ ਸ਼ਾ–ਨਦਾਰ ਹੋ।<br>ur: راشی، آپ میرے لیے بہت حیرت انگیز ہیں۔<br>te: ర సూషి ,నువ్వు నాకు చాల అద్భుతానివి. | Words like ugh, blah, meh convey negativity but leave enough fuzziness for the translator to choose from a range of negative sentiments. |
| 8 | en: liar, liar, pants on fire.<br>hi: झूठे कहीं के।<br>mag: झूठा कहीं के।<br>ml: സത്യസന്തരല്ലാത്ത ആളുകൾ<br>mr: खोटारडा, खोटारडा, खोटे बो–लणारा नंतर त्याच्या खोट्याचा फटका खातो.<br>or: ମିଥ୍ୟାବାଦୀ, ମିଥ୍ୟାବାଦୀ, ନିଆଁରେ ପ୍ୟାଣ୍ଟ\|<br>pa: ਝੂਠ ਦੇ ਪੈਰ ਨਹੀਂ ਹੁੰਦੇ<br>ur: جھوٹا، جھوٹا، پتلون آگ پر۔<br>te: లయ్యర్, లయ్యర్ ప్యాంట్స్ ఆన్ ఫయ్యర్ | en: honest people<br>hi: भरोसे लायक़ लोग हैं।<br>mag: ईमानदार अदमी ।<br>ml: സത്യസന്ധരായ ആളുകൾ<br>mr: प्रामाणिक लोक.<br>or: ସଚ୍ଚୋଟ ବ୍ୟକ୍ତି \|<br>pa: ਇਮਾਨਦਾਰ ਲੋਕ<br>ur: ایماندار لوگ<br>te: నిజాయితీపరులు | "liar, liar, pants on fire." is a poetic proverb which may or may not have a corresponding equivalent in the target language. Here, a creative translator is required. |

Table 14: English (en), Hindi (hi), Magahi (mag), Malayalam (ml), Marathi (mr), Odia (or), Punjabi (pa), Telugu and Urdu (ur) Text Sentiment Transfer Examples (Negative to Positive) (see Section 3.2).

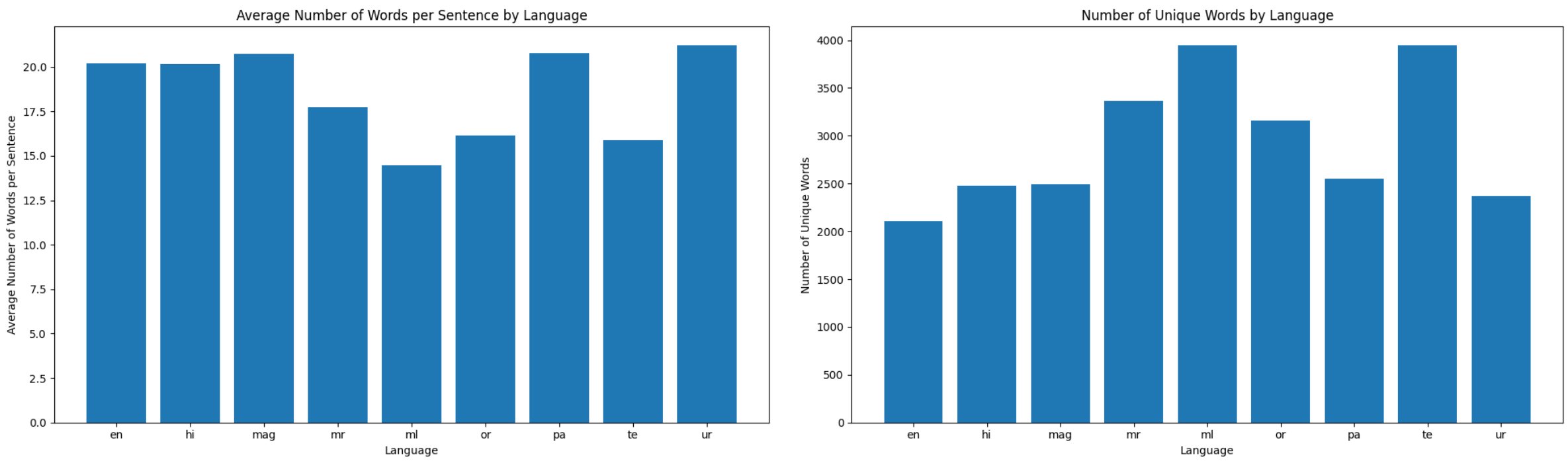

Figure 1: Dataset Statistics: Average number of words per sentence by language (left side), and number of unique words by language (right side)

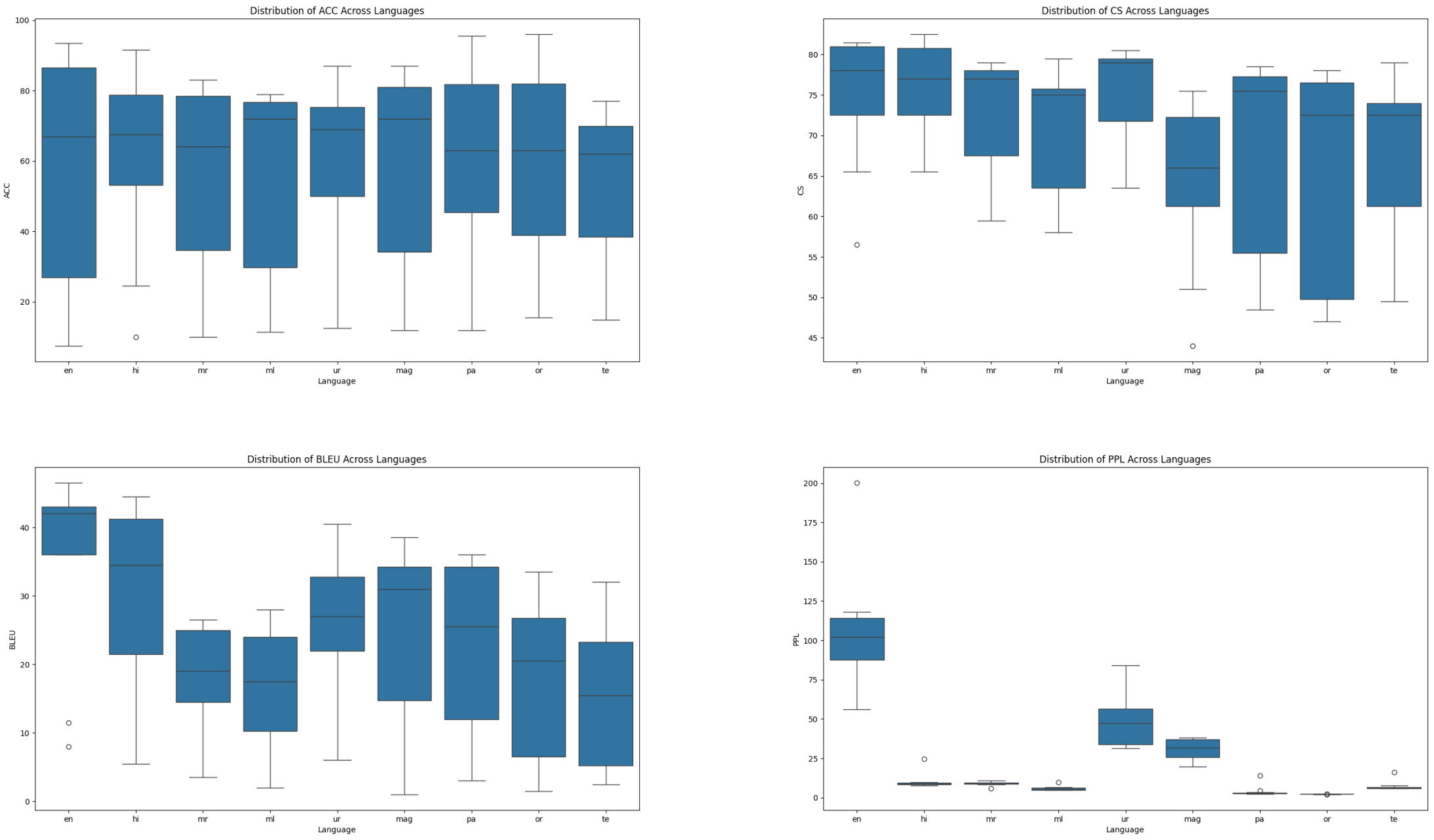

Figure 2: Distribution of ACC, BLEU, CS and PPL across languages respectively

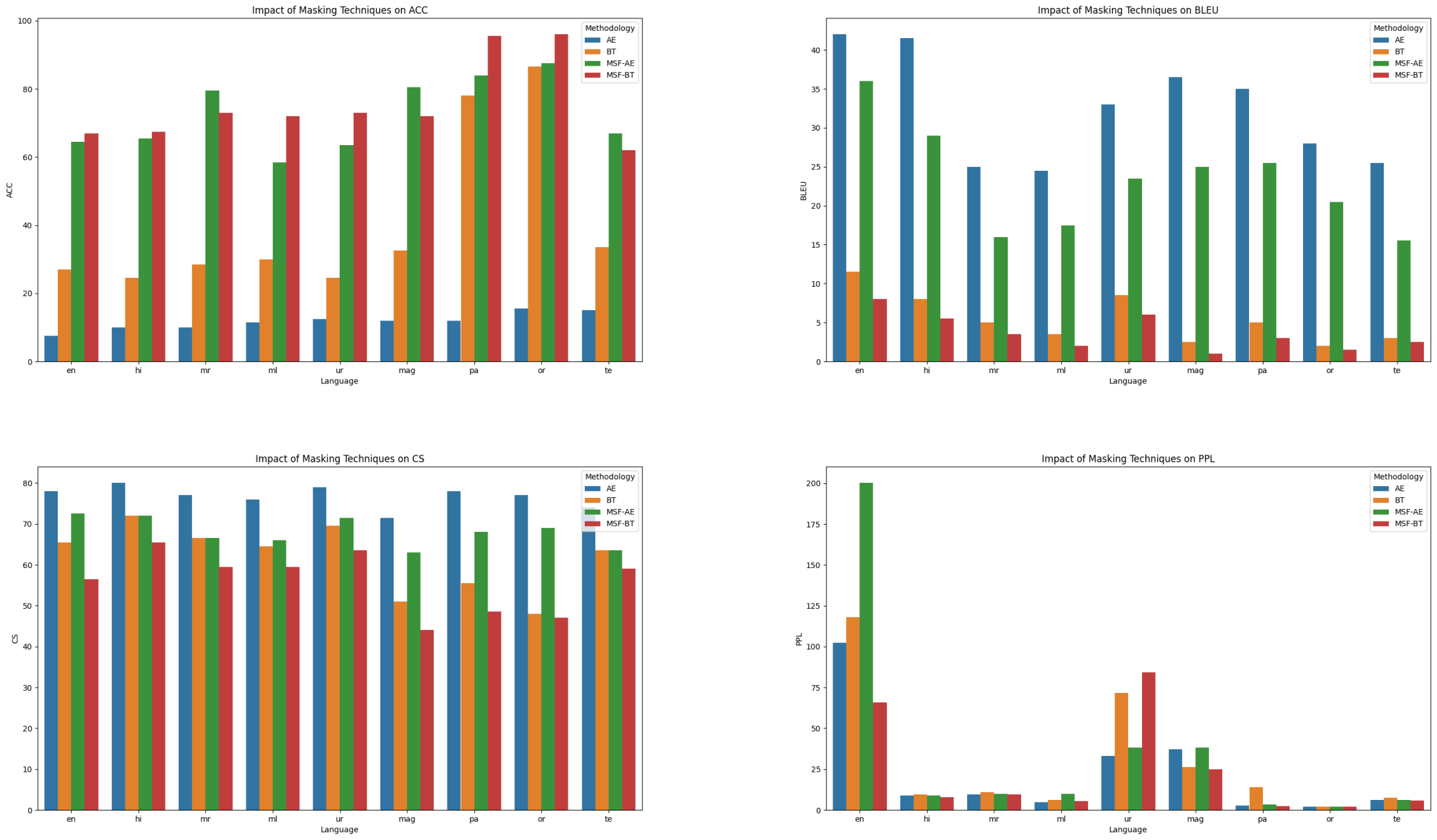

Figure 3: Impact of masking techniques on ACC, BLEU, CS and PPL respectively

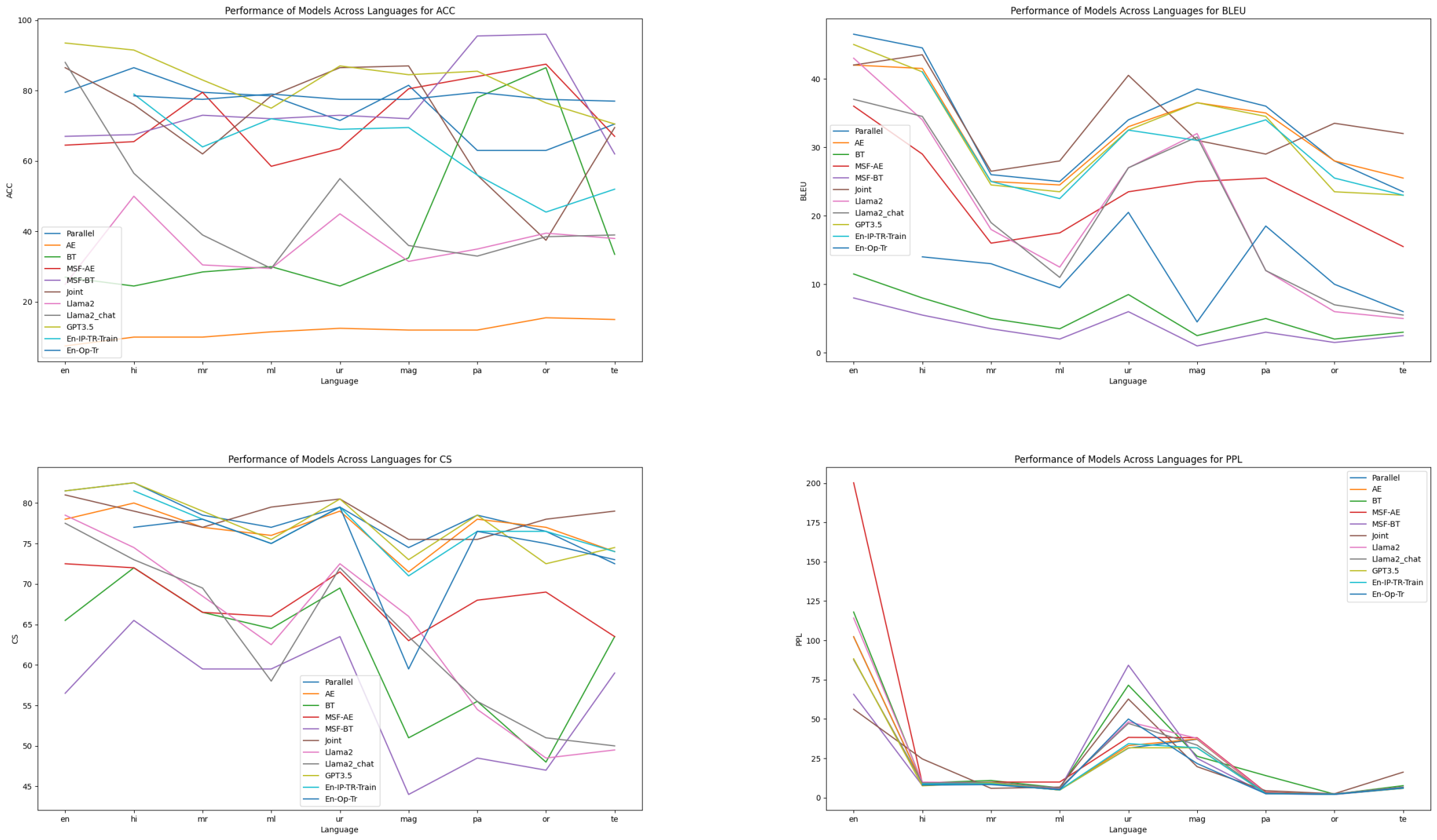

Figure 4: Performance of models across languages for ACC, BLEU, CS and PPL respectively

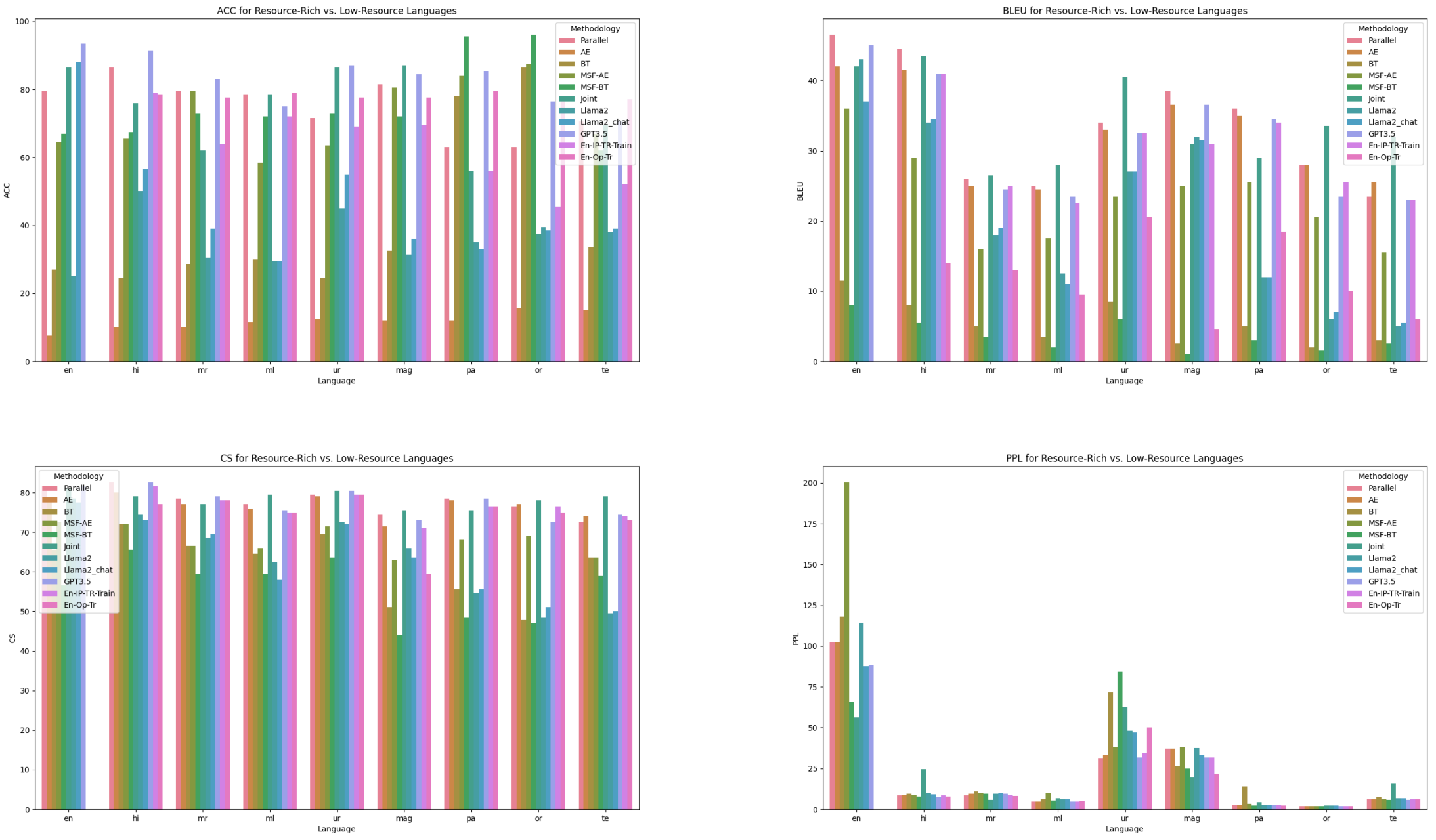

Figure 5: ACC, BLEU, CS and PPL for resource-rich (English and Hindi) vs. other low-resource languages respectively

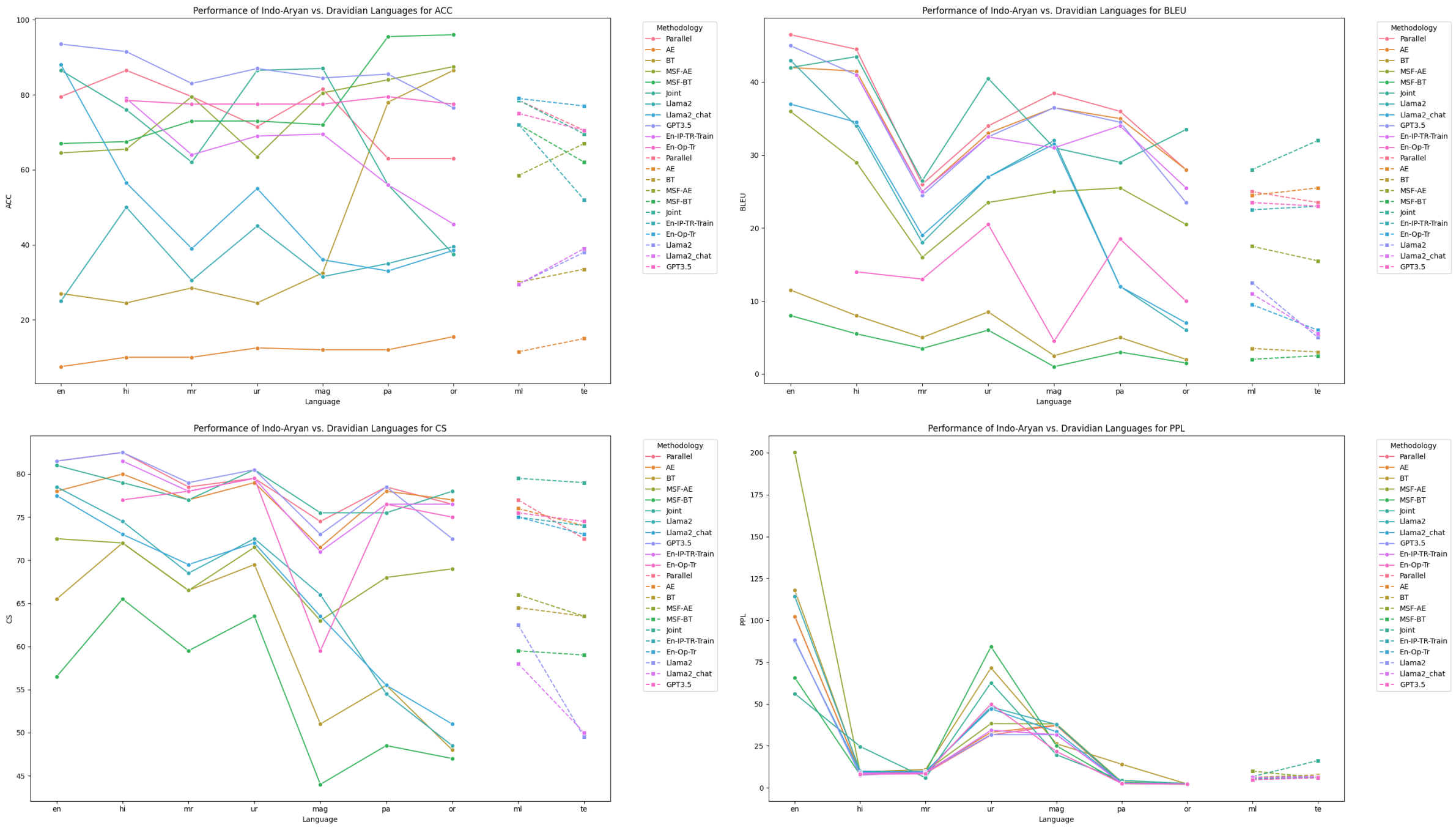

Figure 6: Performance of Indo-Aryan vs. Dravidian languages for ACC, BLEU, CS and PPL respectively

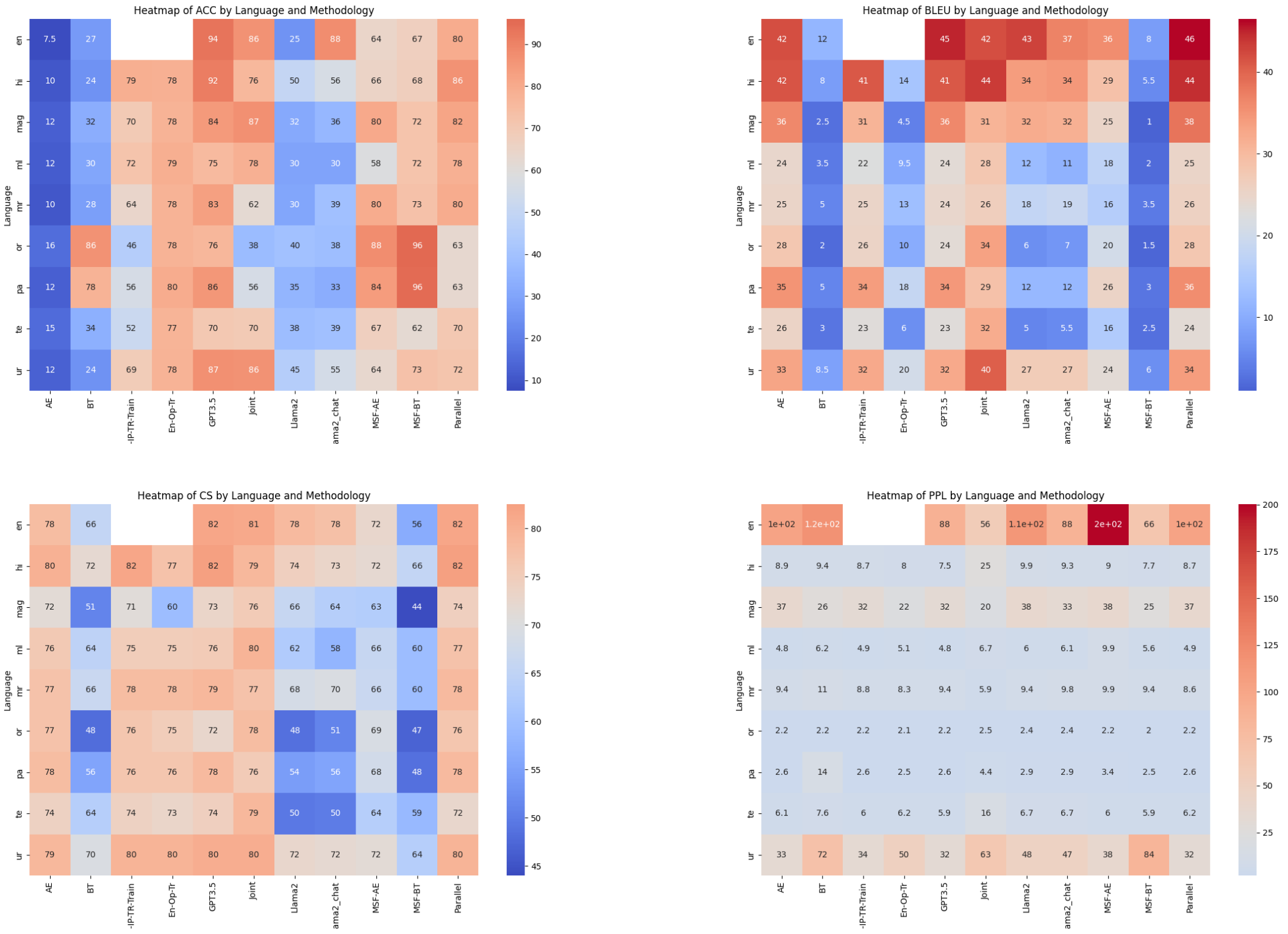

Figure 7: Heatmap of ACC, BLEU, CS and PPL by language and methodology respectively

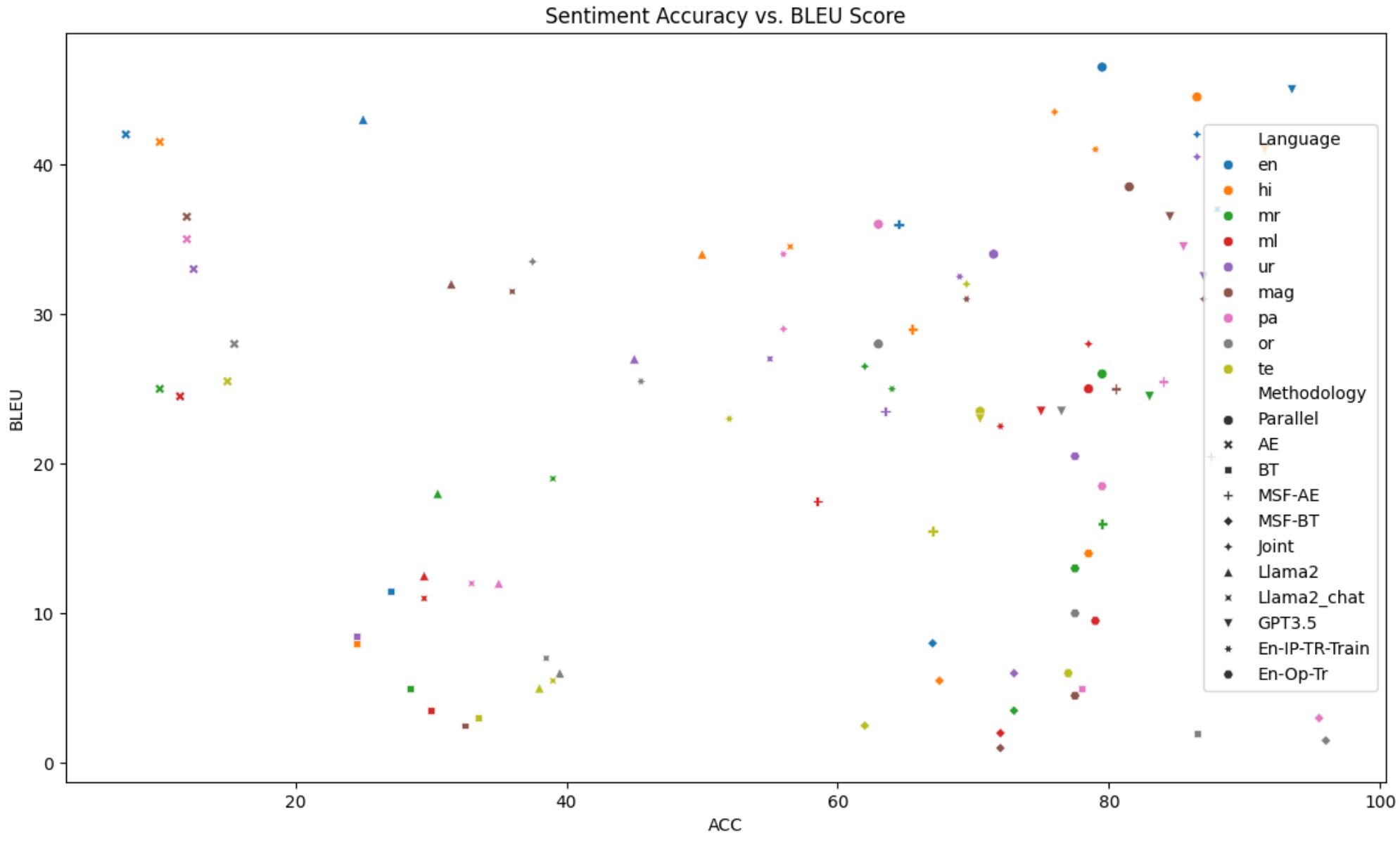

Figure 8: Sentiment accuracy vs. BLEU score across all the languages and models.

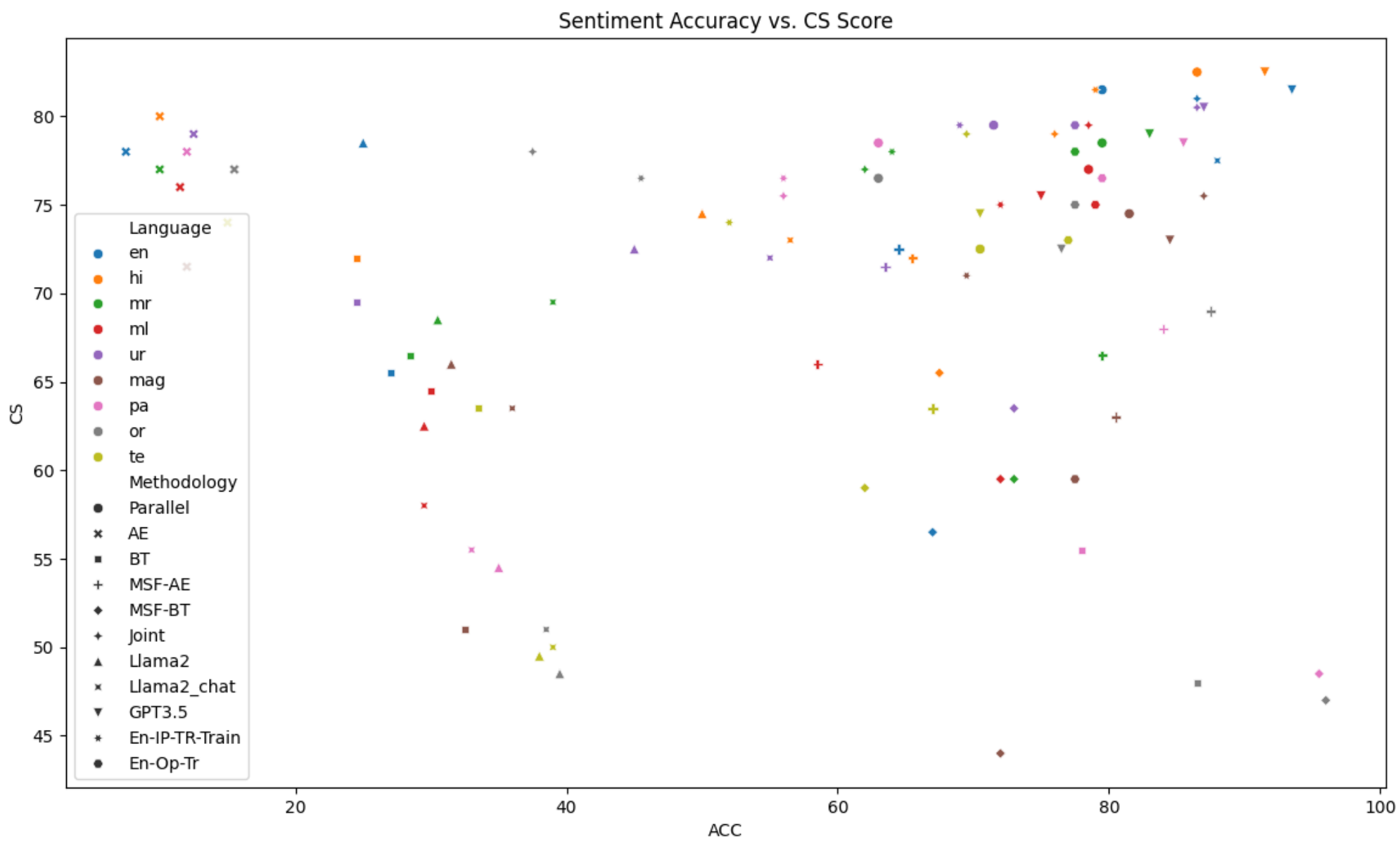

Figure 9: Sentiment accuracy vs. CS score across all the languages and models.

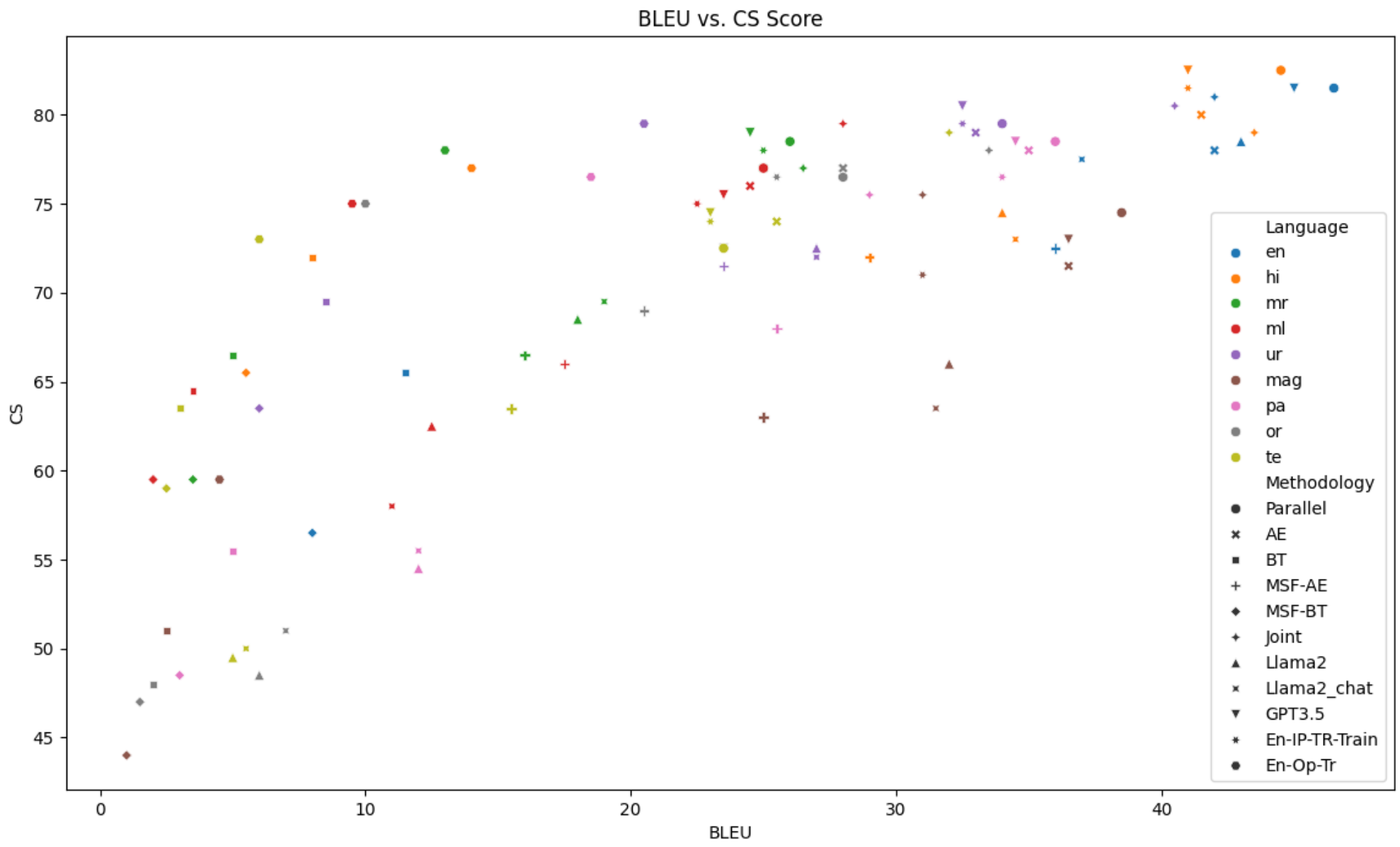

Figure 10: BLEU vs. CS score across all the languages and models.